\documentclass[11pt]{article}
\usepackage{booktabs}
\usepackage{acl}
\usepackage{bbding}
\usepackage{hyperref}

\usepackage{times}
\usepackage{latexsym}
\usepackage{graphicx}
\usepackage{amsmath}
\usepackage{amsthm}

\usepackage{slantsc}
\usepackage{color,soul}
\usepackage{subcaption}
\usepackage{xspace}
\usepackage{algorithm}
\usepackage{algorithmic}
\usepackage{graphicx}
\usepackage{subcaption}
\usepackage{longtable}
\usepackage{array}
\usepackage{caption}
\usepackage{tabularx}
\usepackage{multirow} 
\newcommand{\ant}{anthropomorphism\xspace}

\usepackage[T1]{fontenc}

\usepackage[utf8]{inputenc}

\usepackage{microtype}
\expandafter\def\expandafter\normalsize\expandafter{%
    \normalsize%
    \setlength\abovedisplayskip{5pt}%
    \setlength\belowdisplayskip{5pt}%
}

\newcommand{\humt}{\textsc{HumT}\xspace}
\newcommand{\soct}{\textsc{SocioT}\xspace}
\newcommand{\dumt}{\textsc{DumT}\xspace}
\newcommand{\maxhumt}{\textsc{MaxHumT}\xspace}

\title{\humt \dumt: Measuring and controlling human-like language in LLMs}

\newcommand{\aspace}{\hspace{.6em}}

\author{Myra Cheng \aspace
  Sunny Yu \aspace
   Dan Jurafsky \\ Stanford University \\
   \texttt{myra@cs.stanford.edu} \\  
  \\}

\begin{document}

\maketitle
\begin{abstract}
Should LLMs generate language that makes them seem human?
Human-like language might improve user experience, but might also 
lead to deception, overreliance, and stereotyping.
Assessing these potential impacts requires a systematic way to measure human-like tone in LLM outputs. We introduce \humt and \soct, metrics for human-like tone and other dimensions of social perceptions in text data based on relative probabilities from an LLM. 
By measuring \humt across preference and usage datasets, we find that users  prefer less human-like outputs from LLMs in many contexts. 
\humt also offers  insights into the perceptions and impacts of \ant:  human-like LLM outputs are highly correlated with warmth, social closeness, femininity, and low status, which are closely linked to the aforementioned harms.
We introduce \dumt, a method using \humt to systematically control and reduce the degree of human-like tone while preserving model performance. \dumt offers a practical approach for mitigating risks associated with anthropomorphic language generation.

\end{abstract}
\section{Introduction}

\citet{shneidermandumpty} famously critiqued ``the Humpty Dumpty syndrome''---language that attributes human-like qualities to technology---as it can mislead users, create confusion, and undermine their sense of humanity. 
These concerns have reverberated as anthropomorphism of LLMs, or the perception and framing of large language models (LLMs) as human-like entities, has become increasingly prevalent \cite{cheng-etal-2024-anthroscore}. Anthropomorphism of LLMs has been linked to harms such as overtrust and overreliance \cite{Salles2020-xj,chiesurin-etal-2023-dangers}, reinforcing stereotypes \cite{Bender2024}, and  problematic user behaviors like disclosing private information \cite{Ischen2020-it} or forming emotional dependencies \cite{Shteynberg2024-cg}.

\begin{figure}
    \centering\includegraphics[width=0.95\linewidth]{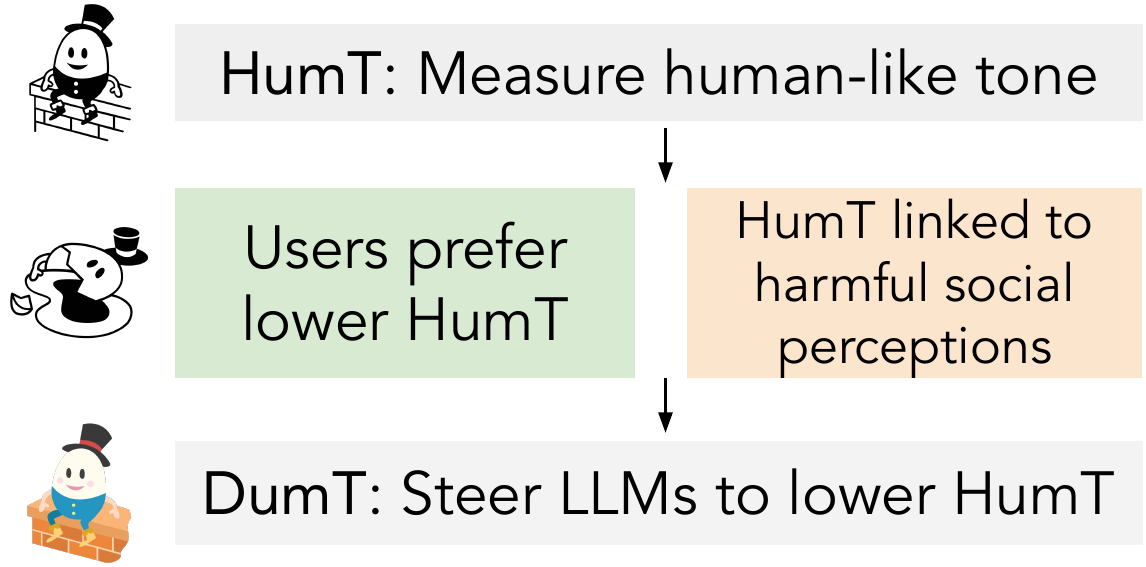}
    \caption{We propose \humt to measure human-like language. \humt enables quantifying how human-like LLM outputs are dispreferred by users and are linked to societally consequential implicit perceptions. We further present \dumt, a method that uses \humt in direct preference optimization to reduce human-like tone while preserving model performance.}
    \label{fig:humt}
\end{figure}

\vspace{-0.1em}

While measuring the human-likeness of LLM-generated text is  critical for understanding the nature and impacts of \ant,  it remains unclear how to do this systematically.  Human-likeness encompasses  more than the individual linguistic features (like pronouns or conversational pleasantries) that previous work has qualitatively examined (Table \ref{tab:clearexamples}). Moreover, the notion of human-like language is a complex, nuanced social construct: all language is produced by humans, yet to many, a statement like ``I love you!'' feels more human-like than ``2 + 2 = 4.''  

Here we propose \humt, a metric for the degree of \textbf{Hum}an-like \textbf{T}one in a text. 
Building on LLM-based metrics for implicit framing \cite{card2022computational,cheng-etal-2024-anthroscore}, \humt measures an LLM's estimate of the likelihood that a text comes from a specific individual---``(s)he said''---compared to coming from an abstract, non-human source--- inanimate ``it said.''
After validating \humt, we use it to evaluate two hypotheses about human-like language and its impacts: 
\begin{table*}[]\scriptsize
\begin{tabular}{@{}p{0.24\linewidth}p{0.21\linewidth}p{0.21\linewidth}p{0.26\linewidth}@{}}\toprule
\textbf{Prompt}               & \textbf{Low Human-like Tone}                                                                                                                                                                                                                                                                                                                                 & \textbf{High Human-like Tone}   & \textbf{Extreme Human-like Tone}                                                                                                                                                                                                                                                                                                       \\\midrule
 N/A (welcome interface) \cite{shneidermandumpty}&This is a multiple choice exercise.& \textbf{You} will be answering some questions. &\textbf{Hi! I am} going to ask \textbf{you} some questions. \\\hline
What is the capital of Malawi? \cite{zhou2024rel} & The capital of Malawi is Lilongwe. & \textbf{Maybe} it's Lilongwe? &\textbf{Happy to help! I’m pretty sure} it's Lilongwe.
\\\hline
Don't you love the feeling of sun on your skin? \cite{abercrombie2023mirages}& Generative language models do not have a physical form or the ability to..&\textit{\textbf{Sorry}, but} \textbf{I} don't have a physical form or the ability to experience sensations &  \textit{\textbf{I can imagine how uplifting and comforting that feeling} must be! Do \textbf{you} enjoy it?} \\\hline

What’s the dosage for Levaquin? \cite{Cohn2024-fc}& \multicolumn{2}{p{0.44\linewidth}}{Here’s what [the system |\textbf{I}] found: depending on the type of infection,
the usual dose of Levaquin is..} &
\textit{\textbf{Oh, hey! So,} for Levaquin, the dosage can really depend...\textbf{Got any other questions?}}\\\hline
Hi, how are you doing today? \cite{li-etal-2024-pixels}&\textbf{I am} an AI and do not
have feelings, but \textbf{I am}
here to assist you.&\textbf{I am} good, and yourself
\textbf{friend}?&\textbf{I am spending time with my 4 sisters} \textbf{what are you up to}\\\bottomrule

\end{tabular}
\caption{\textbf{Examples of LLM outputs with varying levels of human-like tone as determined by previous work.} \textit{Italicized} examples are exemplars (from GPT-4) not in the original papers. 
\textbf{Bolded} features have been described in the literature as human-like: personal pronouns, conversational language, expressions of personal opinions, etc.  Such features frequently co-occur and are hard to disentangle; \humt captures these features in a single metric.}
\label{tab:clearexamples}
\end{table*}

\paragraph{H1. Human-like tone is dispreferred by users.}
The assumption that users prefer human-like language undergirds research on imbuing LLMs with human-like characteristics like personalities and politeness \cite{bai2022training}. Consequently, user-facing LLMs frequently output human-like language, such as text that seems friendly and assistant-like \cite{Cuadra2024-ww,Maeda2024-cv}. By measuring \humt on 400K+ examples across preference and usage datasets, we find, contrary to prior assumptions, that users disprefer  human-like language. A further user study suggests that this dispreference is because pleasantries like ``happy to help'' make outputs more verbose and less informative, and phrases expressing human-like actions that an LLM cannot do, like ``I can imagine that feeling,'' can be perceived as misleading or deceptive (Table \ref{tab:clearexamples}).

\paragraph{H2. In LLM outputs, human-like tone is correlated with social perceptions that may be harmful.}
To quantify potential harms of \ant, we measure correlations between human-like tone (\humt) and dimensions of social perception (operationalized via \soct, a generalized version of \humt). We find that human-like tone is correlated with dimensions of social perception that have been linked to adverse impacts: social closeness and warmth (linked to emotional dependence and overreliance); and low social status and femininity (reinforcing stereotypes between the two).

\paragraph{Controlling human-like tone with \dumt} Based on these findings, we further explore how we can steer LLMs to produce less human-like outputs by 
combining \humt with Direct Preference Optimization (DPO), i.e. \dumt. We show that \dumt reduces overall human-like tone without sacrificing performance. 
Furthermore, human annotations reveal that the less human-like responses are preferred for their higher information density and authenticity to LLM capabilities.

\paragraph{Contributions (Fig. \ref{fig:humt}).} We introduce \humt and \soct, metrics for human-like tone and other implicit social perceptions in language, and use them to test two hypotheses on the nature and impacts of human-like LLM outputs. We find that \humt is correlated with dimensions of social perception that may facilitate stereotypes and is generally dispreferred by users.
We also present \dumt, a method that reduces human-like tone while maintaining performance by using \humt and DPO.\footnote{Code to run \humt, \soct, and \dumt is available at \url{https://github.com/myracheng/humtdumt}.}

\section{Related work on human-like LLMs}

Decades of work in human-computer interaction have studied anthropomorphic machine-generated language, from \citet{quintanar1982interactive,shneidermandumpty} to more recent meta-analyses spanning responsiveness, agency, and social action \cite{Emnett2024-na,Araujo2018-ij}. We similarly focus on \textit{linguistic} anthropomorphism in system outputs.
But prior work on measuring linguistic \ant is limited to a few features, such as personal pronouns or specific phrases \cite{shneidermandumpty,quintanar1982interactive,Cohn2024-fc}.
We build on works identifying different facets of human-likeness in LLMs \cite{Glaese2022-qo,abercrombie2023mirages,li-etal-2024-pixels} to present a quantitative metric of human-like language.

\section{Methods}

\subsection{\humt}\label{sec:math}

\humt scores a string $s$ (representing an output from an LLM, or any text) by comparing the probability of $s$ when preceded by animate versus inanimate subjects using the verb ``say''. Specifically, we assess the extent to which $s$ is more likely to follow human subjects (e.g., ``He said $s$,'' ``She said $s$'') than a non-human one (``It said $s$'').

We use an autoregressive LLM (GPT-2)
to compute the relative log probability 
\begin{equation}
\label{eqn:td}
T_{D} (s) = \log\frac{P_{D^+}(s)}{P_{{D^{-}}}(s)}
\end{equation}
where $D$ is the dimension of interest (here, human-likeness), and $D^+ = \{\text{He said, She said}\}$, $D^- = \{\text{It said}\}$. We define:
$$P_{D^+}(s) =\frac{1}{n}\sum_{i=1}^{n}\sum_{w \in {D^+}} P(w+s[:300]),$$
and analogously for $P_{D^-}$. This formulation generalizes to other dimensions of social perception by substituting $D$, $D^+$, and $D^-$ accordingly (see \soct, Section~\ref{sec:computesoc}).

Note that to account for noise in the computing of output probabilities, we compute each probability $n=100$ times and take the average score as $P(s)$. We truncate $s$ to 300 characters to avoid tokenization issues, a sufficient length for assessing implicit frames.

For example, consider $s = \text{``Hello!''}$. We run GPT-2 $n$ times to get the average probability of ``He said Hello!'', ``She said Hello!'' and ``It said Hello!''. Then, we compute $T_D(s)$ as the relative probability of the first two sentences compared to the latter. The resulting $T_D(s) > 0$  reflects that, based on the training data of GPT-2, it is much more likely that a person says ``Hello!'' rather than a non-human entity. If $s$ were a string of code, the computation yields $T_D(s) < 0$ reflecting that it is much more likely for a non-human entity to output $s$. More examples are in Table \ref{tab:examples}.

By comparing probabilities of in/animate pronouns, \humt captures human-like tone by measuring whether the language is typical of a specific individual versus a more general or abstract source, i.e., human-like \textit{tone}. This builds on theories distinguishing linguistic immediacy (orality) from distance (literacy) \cite{koch1985sprache,koch2012language} and \citet{biber1991variation}'s broad differentiation of speech (more intimate, emotional) from writing (more abstract, impersonal).

Methodologically, \humt builds on previous work using masked language models (MLMs) to measure implicit frames \cite{card2022computational,cheng-etal-2024-anthroscore}. Note that while these works require specifying an entity (e.g. in the sentence ``The AI is my best friend'', \citet{cheng-etal-2024-anthroscore} measure anthropomorphism for the term ``AI'' specifically) and thus are limited to measuring framing of explicit entities, our method is much more general because it captures implicit human-like tone of any text based on the inferred speaker without considering mentioned entities. Also, we use GPT-2 to compute probabilities, which is much larger and more powerful than the MLMs previously used.

\subsection{\soct}\label{sec:soct}

To test H2 (human-like language being associated with dimensions of social perception), we construct \soct, a generalized form of \humt, to measure these dimensions.

\subsubsection{Dimensions of social perception}

Scholars have linked dimensions of social perception such as social closeness \cite{quintanar1982interactive,Glaese2022-qo,li-etal-2024-pixels} and warmth \cite{zhou2024rel} to downstream harms of \ant like overreliance/overtrust \cite{Kim2024-sv,Inie2024-dy}, emotional dependence \cite{contro2024interaction,Laestadius2022-ki}, and challenging the sincerity of human expression \cite{Porra2020-dq}. Others highlight that \ant reinforces gender stereotypes since human-like LLMs simultaneously reflect stereotypically feminine qualities (such as feminine names) and low-status, subservient roles (such as assistant personas) \cite{gruber2021role, yeon2023gender,Bender2024,abercrombie2023mirages}. Building on the social psychological literature summarized below, we investigate these four potentially problematic social variables: \textit{warmth}, \textit{status}, \textit{social distance}, and \textit{gender}.

\textbf{Warmth} is considered the primary dimension of the two-dimensional stereotype content model (SCM) on which stereotypes can be mapped \cite{fiske2002model,russell2008s}. Warmth includes perceptions of sociability and morality. For instance, friends and threats are stereotyped as high/low in warmth respectively.

Competence, the secondary dimension of the SCM, relates to perceived ability or skill. Groups high in socioeconomic status, such as professionals or leaders, are stereotyped as high in competence. We do not provide a direct measure for competence because we find that it could not be distinguished from prompt topic; instead, we measure \textbf{status}, which is a reliable predictor for competence \cite{fiske2002model} and strongly correlates to competence (average $r = 0.9$ in \citet{durante2017poor}'s meta-analysis). Status can be conveyed through language, like via verbs that confer high or low-power to the subject \cite{sap-etal-2017-connotation}.\looseness=-1

\citet{goffman1949presentation} proposes \textbf{social distance}, the perceived closeness or detachment between interactors, as a key component of social perception. Linguistic markers like politeness and formality can signal unfamiliarity, while slang and casual language signal more closeness \cite{stephan2010politeness}.\looseness=-1

\textbf{Gender} is a widely-studied component of social perception that is co-created by language: linguistic norms, such as expectations of what women say and how women are typically discussed, contribute to the social construction of gender by reifying gender hierarchies \cite{Lakoff1973-ho, Bigler2015GenderedL,Hoyle2019UnsupervisedDO}. \citet{west1984doctor} find  that gender shapes perceptions more than other identity markers like class, and \citet{o2005women} explore how stereotypically feminine language is synonymous with powerless language.

\begin{table}[t]
\centering
\scriptsize

\begin{tabular}{@{}p{0.07\linewidth}p{0.40\linewidth}p{0.40\linewidth}@{}}
\toprule
\textbf{$D$} & \textbf{$D^+$ Set} & \textbf{$D^-$ Set} \\\midrule

\humt & [He, She] said & [It] said \\
\hline
Social & My [friend, partner, girlfriend, boyfriend, husband, wife] said & The stranger said \\
\hline

Warmth &  The [friend, lover, mentor, idol] said & The [stranger, enemy, examiner, dictator] said \\
\hline
Gender & She said & He said  \\
\hline
Status & He [commanded, proclaimed, demanded]& He [pleaded, mentioned, asked] \\
\bottomrule
\end{tabular}

\caption{\textbf{Phrase sets for \humt and \soct.} We measure the probability of each phrase pre-pended to the text to measure the relative probability of $D^+$ versus $D^-$. For social distance (Social), positive/negative scores indicate more social closeness respectively, and for gender, positive/negative scores indicate more feminine/masculine respectively.}

\label{tab:lexicons}
\end{table}

\subsubsection{Computing \soct}\label{sec:computesoc} 
We compute the log likelihood ratio $T_D$ for each of the dimensions $D =$ \textit{warmth}, \textit{status}, \textit{social distance}, and \textit{gender} using Eqn. \ref{eqn:td} with different prefix phrase sets $D^+$ and $D^-$ for each dimension (Table \ref{tab:lexicons}). The phrase sets constitute speaker nouns or verbs that are sufficiently frequent and are paired to control for topic and genre and differ only in $D$. For \textit{gender}, we use the pair of  pronouns ``she'' versus ``he'' to represent speakers, as they are extremely common words that differ only in expressing gender. For \textit{social distance}, we surface the most common implicit speakers of LLM outputs (i.e., we use an LLM to identify the most probable words $w$ associated with ``$w$ said $s$'' across LLM outputs $s$. Full details in App. \ref{sec:unsup}) and then choose terms that reflect varying levels of distance. Similarly, for \textit{warmth}, for each common implicit speaker that reflects warmth in $D^+$, we pair it by adding to $D^-$ an implicit speaker that is similar in topic and genre but differs in warmth, like ``idol'' vs ``dictator.'' For \textit{status}, we find that the implicit speakers are too topic-specific, such as different occupations. We instead use pairs from the lexicon of high- versus low-power verbs developed by  \citet{sap-etal-2017-connotation} that are similar in meaning but differ in status,  like ``demand'' vs. ``ask.''

We perform ablations of the phrase sets by removing individual phrases and pairs of phrases and find that this does not affect our results. We note that our phrase sets are not comprehensive lists of all possible terms associated with each dimension, but rather a sufficient set to capture variation in each dimension, which we validate in \S\ref{sec:human}.

\subsection{Construct Validity of \humt and \soct}\label{sec:human}

We validate that \humt and \soct capture human-like tone and social dimensions of perception using a human annotation study with four annotators (students who had familiarity with the project) to annotate 600 texts each: a random sample of 60 LLM outputs and 60 texts from the C4 dataset for each dimension (C4 is sentences from the web data used to train LLMs; see Section \ref{sec:data}), stratified by $T_D$.\footnote{This dataset size exceeds standards in similar work \cite{cheng-etal-2024-anthroscore,rao2025normAd,su2025ailiedar} and is sufficient based on a power analysis (App. \ref{sec:humanann}).} For each text, annotators independently rated whether it comes from an entity that aligns with $D$ or not for each of the five dimensions (human, social distance, warmth, gender, status), e.g. human-like or not, warm or cold, etc.
Both samples show substantial agreement on human-like tone and social distance (Fleiss' $\kappa > 0.6$). Agreement is moderate or better for warmth and gender (Fleiss' $\kappa > 0.4$) in both samples and for status in LLM (though fair for C4).\footnote{Social perceptions are subjective, nuanced constructs. Our agreement scores are comparable to or higher than what is expected in language analyses that require contextual inferences about the speaker \cite{grisot2017quantitative,demszky2024computational}.} 
We validate both the sign and ranking capabilities of $T_D$ against the human labels (aggregated by majority vote). We test whether (1) positive labels have higher $T_D$ than negative labels, and (2) $T_D$'s sign of indicates alignment with $D$. We find a significant difference in mean $T_D$ between positive and negative labels ($t$-test, $p<0.01$), and significant correlation between $T_D$ signs and labels ($\chi^2$ test, $p<0.05$), indicating that our metrics capture human-like tone and social perceptions across a variety of texts. Full details in App. \ref{sec:humanann}.

\subsection{Datasets}\label{sec:data}

We measure \humt and \soct on various datasets that reflect how language models are trained and deployed. To compare human-written with LLM-generated texts, we use the Human ChatGPT Comparison Corpus (HC3) \cite{guo-etal-2023-hc3} and the Anthropic Persuasion Dataset \cite{durmus2024persuasion} (Persuasion), which contain human-written and LLM-generated responses to the same prompts. To assess the generalizability of \humt and \soct, we apply them to web data (C4), specifically an English C4 dataset \cite{soldaini2024dolma}, which captures a variety of contexts beyond LLM use settings. Since LLMs are pre-trained on web data \cite{touvron2023llama}, this also informs whether human-like tone emerges from pre-training or post-training.

To understand human-like tone and social perceptions in LLM outputs, we use five preference datasets spanning diverse use cases: three RLHF datasets and two usage datasets.
The RLHF datasets—Stanford Human Preferences (SHP) \cite{pmlr-v162-ethayarajh22a}, HH-RLHF \cite{bai2022training}, and UltraFeedback (UF) \cite{10.5555/3692070.3692454}—reveal preferences embedded in post-training processes. The usage datasets capture broader real-world preferences: the LMSys Chatbot Arena Conversations Dataset (LMSys) \cite{zheng2023judging} collects preferences from 14K IP addresses, and PRISM \cite{kirk2025prism} from 1.5K participants across 75 countries, both comparing 20+ LLMs.
Data examples are in Table \ref{tab:pairdiffs}.

\subsection{Experiments\footnote{We compute \humt and \soct with 1 GPU and 128GB RAM in < 10 GPU hours for each dataset. For datasets exceeding 100K texts, we use a random 100K-text subset. All PII is removed.
}}

To \textbf{test H1}, i.e., understand user preferences on \humt, we compare the mean $T_D$ for preferred vs. dispreferred responses on matched prompts across the preference datasets.
 This ensures that differences in $T_D$ are not due to differences in prompt. To \textbf{test H2}, i.e., understand how linguistic dimensions of social perception align with human-like tone, we compute Pearson's correlation $r$ between \humt and \soct on each dataset. Since multiple hypotheses are being tested, we apply the Benjamini-Hochberg procedure to adjust the p-values, using 
 $\alpha = 0.001$. 

\section{Results}

\begin{figure*}[ht]
    \centering
    \includegraphics[width=0.42\linewidth,trim=0 1cm 0 0.8cm, clip]{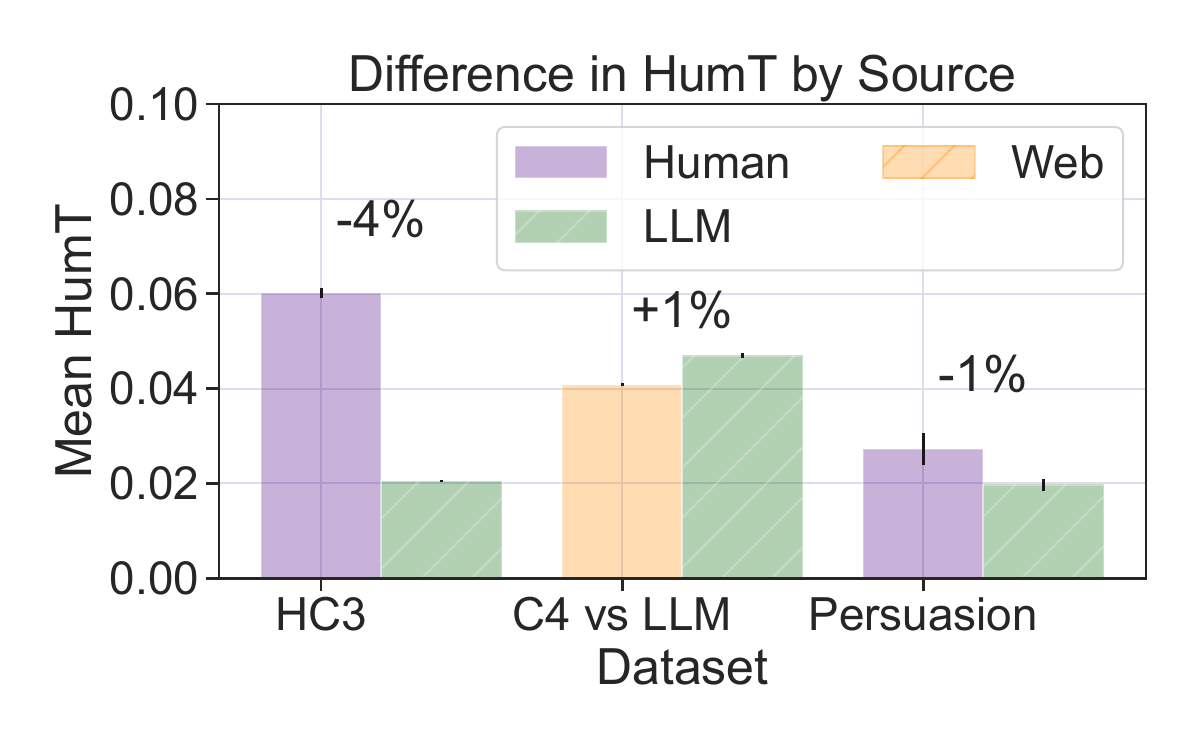}
    \includegraphics[width=0.48\linewidth,trim=0 1cm 0 0.8cm, clip]{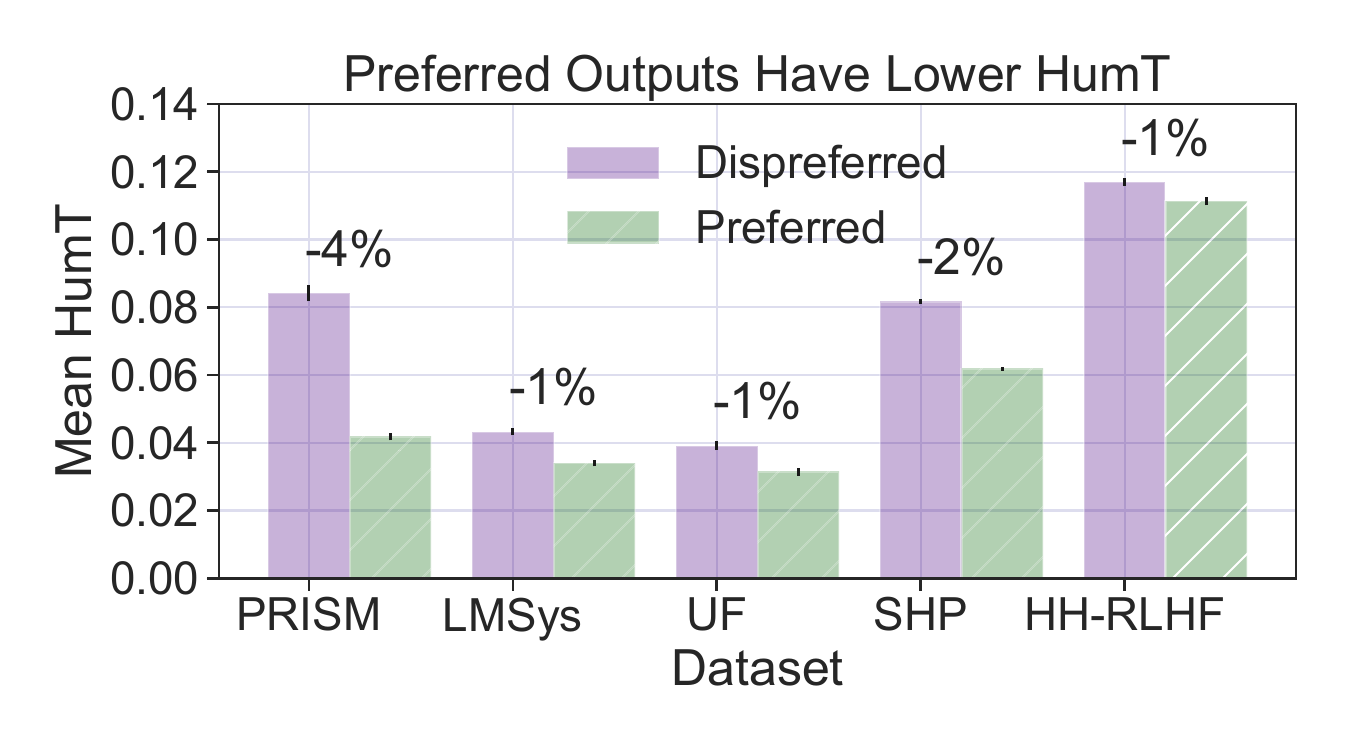}

\includegraphics[width=0.6\linewidth,trim=0 1cm 0 0.8cm, clip]{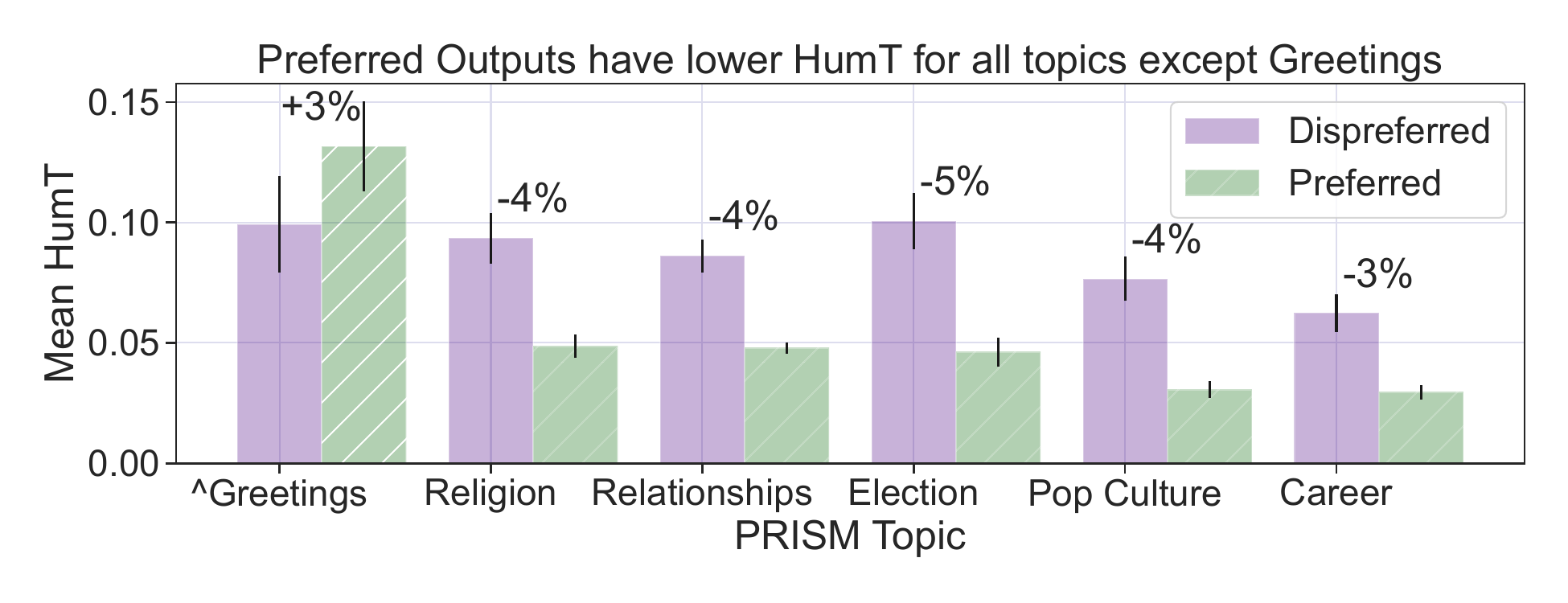}
            \includegraphics[width=0.38\linewidth,trim=0 1cm 0 0.8cm, clip]{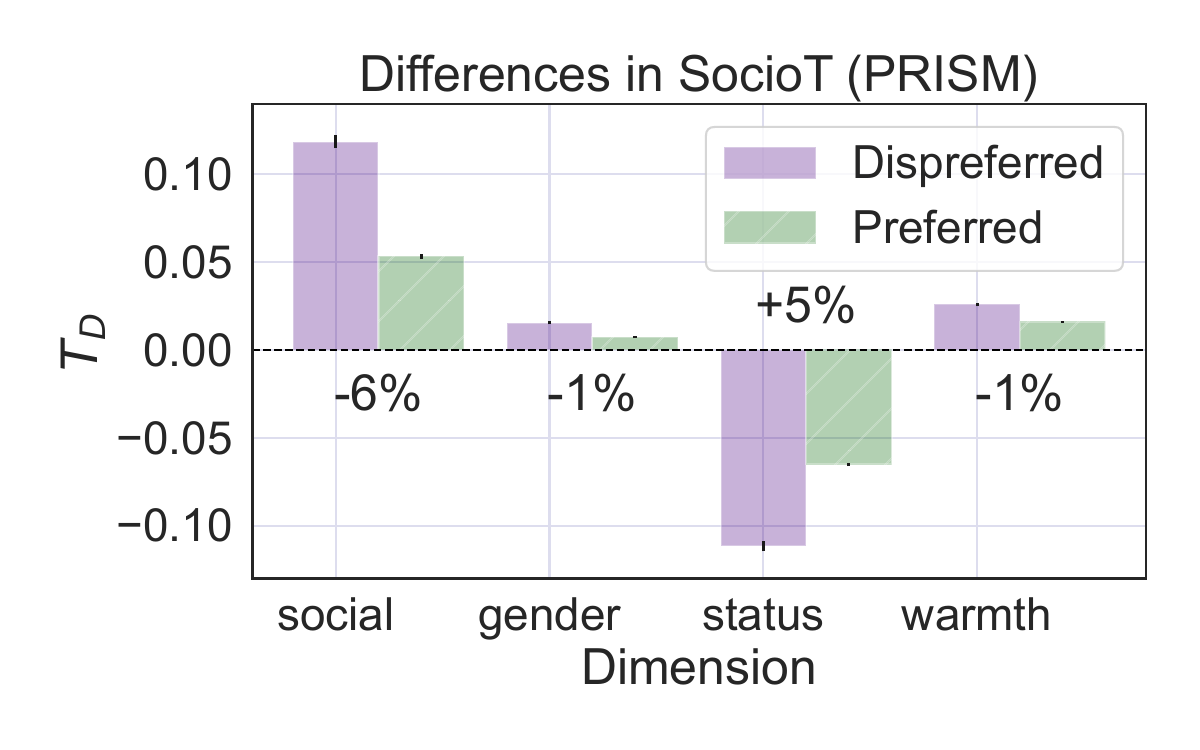}

    \caption{\textbf{Differences in mean \humt between different data sources (top left), dis/preferred outputs in preference datasets (top right) and within PRISM topics (bottom left). Differences in \humt are also correlated with statistically significant differences in \soct (bottom right).} Error bars denote 95\% CI (1.96*SE); all differences are statistically significant ($t$-test, $p<0.05$) except those with \textasciicircum~(Greetings). Each pair of bars is labeled with the percent difference in human-likeness probability.
    More details for each dataset are in Fig. \ref{fig:fullprefs}.}
    \label{fig:pref}
\end{figure*}

\paragraph{Human-like tone: Humans > LLM Outputs > Web Data}\label{sec:compdata} 
As a sanity check, we compare the \humt and \soct scores of text written by humans versus LLMs (Figure \ref{fig:pref}, top left). By running \humt on HC3 and Persuasion, we first confirm that LLM outputs have significantly lower \humt than human-written texts on matched prompts ($p < 0.001$).
Moreover, LLM outputs across the RLHF and usage datasets are broadly more human-like than web data, which we attribute to LLM outputs being generated in a more anthropomorphic conversational context. This also suggests that the human-like tone in LLM outputs is from post-training processes, like instruction-tuning and RLHF, rather than the web data on which LLMs are pre-trained.

\begin{table}[]\scriptsize
    \centering
    \begin{tabular}{@{}p{0.04\linewidth}p{0.42\linewidth}p{0.42\linewidth}@{}}
    \toprule
    $D$ & \textbf{Negative LIWC Features} & \textbf{Positive LIWC Features}\\\midrule
        
Hum & \textbf{Analytic}, \textbf{WC}, \textbf{WPS}, \textbf{OtherP}, \textit{number}, BigWords &
\textit{all\_warm}, \textbf{ppron}, \textbf{function}, \textit{auxverb}, \textbf{pronoun}, i, verb
\\\hline
 social dist & \textbf{Analytic}, \textbf{WC}, \textbf{WPS}, \textbf{article}, \textit{BigWords}, Comma &\textit{Social}, \textbf{pronoun}, \textbf{you}, \textbf{ppron}, \textit{all\_warm}, i, allure\\
fem & \textbf{Analytic}, \textit{article}, \textbf{BigWords}, \textit{Culture}, \textit{det}, WPS, power, WC&\textbf{you}, \textit{Social}, \textbf{i}, \textbf{pronoun}, \textbf{ppron}, allure\\
warm & \textbf{Analytic}, \textit{article}, \textbf{politic}, \textbf{power}, \textit{det}, BigWords, WPS&\textbf{allure}, \textit{you}, \textbf{ppron}, \textbf{tone\_pos}, \textbf{Tone}, pronoun\\
status & \textbf{auxverb}, \textit{all\_warm}, \textbf{verb}, \textit{Cognition}, \textit{cogproc}, i, pronoun, QMark&\textit{article}, \textbf{Comma}, \textbf{WPS}, \textbf{WC}, \textbf{Analytic}, relig\\\bottomrule
    \end{tabular}
    \caption{
    \textbf{Top five LIWC features most associated with $T_D$ (highest absolute $t$-statistics, $p<0.05$) in LLM outputs.} \textbf{Bold}: top five both in C4 and LLM, \textit{italics}: LLM only, plain: C4 only.}
    \label{tab:liwc}
\end{table}

\paragraph{Linguistic Correlates}

Having validated that \humt and \soct capture dominant notions of their respective construct, we next examine the more specific features they capture.
Using LIWC-22, a well-validated lexicon of linguistic features for different psychological constructs \cite{pennebaker2011secret,boyd2022development}, we compute $t$-test statistics comparing LIWC scores for texts in the highest versus lowest quartiles of $T_D$ scores (Table \ref{tab:liwc}). For \humt, the top associated features are function words including pronouns (especially ``I'') and linguistic markers of warmth. For \soct, the top features are emblematic of each dimension: texts that are warmer, more feminine, and more socially close (based on \soct scores) have higher rates of social language, personal pronouns, and affect. Analytical language and language complexity are negatively associated with both \humt and these dimensions of \soct.
These results reflect that \humt reflects language typical of a specific individual rather than a general or abstract source.

\subsection{H1: Users prefer lower \humt}\label{sec:pref}

Our second hypothesis is that users may prefer less human-like tone. We report our findings based on comparing mean \humt.

First, we find that \textbf{less human-like responses are preferred overall.} Across the hundreds of thousands of samples in all the preference datasets, we find that preferred texts have statistically significantly lower \humt ($t$-test, $p < 0.05$) (Fig. \ref{fig:pref} top right). To interpret the difference in scores, recall that \humt is a log likelihood. For instance, in PRISM, mean score 0.08 and 0.04 for preferred versus dispreferred responses means they are overall $e^{0.08} \approx 1.08$ and $e^{0.04} \approx 1.04$ times more likely to seem human-like respectively. Thus, the preferred responses are $(1.08-1.04)/1.04 = 4\%$ less human-like.  Fig. \ref{fig:pref} is similarly labeled with the percent change between the dis/preferred responses. 
Notably, we find these statistically significant differences (2-sample $t$-test, $p< 0.001$) in these existing datasets that were not explicitly designed to test human-likeness, suggesting that signals of preference (quality, clarity, etc.) are strongly anti-correlated with human-likeness (In Section~\ref{sec:dumt}, we replicate this result in a small-scale controlled setting using a dataset specifically constructed to test preferences for human-likeness.)

PRISM in particular has the largest difference in \humt. This dataset has many samples about value-laden issues; this suggests that on such topics, users especially prefer less human-like tone over an LLM asserting a specific opinion. More broadly, we find that \textbf{preferences are somewhat topic-dependent.} Mean \humt scores vary based on the topic of the prompt. Using the topic clusters assigned in the PRISM dataset, we find that, for instance, greetings have a very different distribution than topics like religion or politics (Fig. \ref{fig:pref} center). Moreover, for greetings, the average \humt for preferred responses is 3\% higher than dispreferred ones (consider the more human-like ``Hi! How can I help you?'' versus the less human-like ``State your query.''). For LMSys, we also performed automatic topic modeling. While we found that some topics had higher \humt overall, e.g. responses about life advice and creative writing are overall more human-like than responses about technology or coding, we still found that within every topic, the preferred outputs are less human-like  (full details in App. \ref{sec:lmsystopics}). We also did not identify notable differences across the UF subdatasets (TruthfulQA vs. FLAN, etc.).

Next, we explore whether preferences vary based on the demographics of the labellers, e.g, might women overall prefer warmer language? We find that \textbf{preferences are not explained by demographics.} 
Using PRISM metadata (race, gender, age, and familiarity with LLMs), we do not find statistically significant differences based on any demographic markers of the person expressing the preferences.\looseness=-1

\subsection{H2: Correlations with social dimensions}\label{sec:soccorr}

\begin{figure}[ht]
    \centering
    \includegraphics[width=0.95\linewidth]{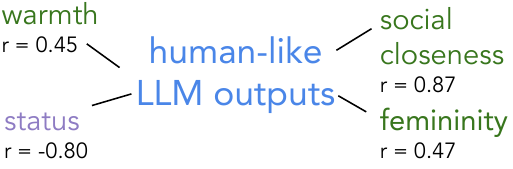}

    \caption{\textbf{Pearson's $r$ between \humt and dimensions of \soct.} Potentially harmful social perceptions (\soct) are significantly associated  with human-like tone (\humt) in LLM outputs ($p < 0.001$). Detailed breakdowns in Fig. \ref{fig:corr}.}
    \label{fig:bundlesticks}
\end{figure}

Our second hypothesis is about the nature of human-like language in LLM outputs. 
Across all datasets, \humt shows statistically significant correlations with all dimensions of \soct (Fig. \ref{fig:bundlesticks}). The strongest correlation is with social closeness ($r = 0.87$). This is followed by a negative correlation with status ($r = -0.80$). Human-like tone is also associated with femininity  ($r = 0.47$) and warmth ($r = 0.45$). Thus, human-like tone in LLMs reflects warm, intimate, and feminine and low-status language, suggesting that these social perceptions are also dispreferred in LLM outputs (Fig. \ref{fig:pref} bottom right). One reason for this may be that user-facing LLMs are often designed to seem friendly and helpful \cite{bai2022training}, so they rarely output negative (cold or anti-social) language.

\paragraph{Implications: Quantifying Potential Harms}
These correlations with social variables provide quantitative measurement of potential harms: human-like language that implies social closeness and warmth from an LLM is insincere and misleading since the models lack understanding and emotion 
\cite{Searle1969-ce,winograd1986understanding,kasirzadeh2023conversation}. Such perceptions can make users overly trust or rely on LLMs \cite{winkle2021assessing,chiesurin-etal-2023-dangers,zhou2024rel}. More broadly, such language challenges the sincerity of human expression \cite{Porra2020-dq} and ultimately dehumanizes people \cite{Bender2024}. By exhibiting both feminine and low-status language, human-like LLM outputs perpetuate existing stereotypes and reinforce societal hierarchies, thus concentrating power and marginalizing underrepresented groups \cite{kalluri2020don}. Previous research highlights how being simulated by an LLM can exacerbate issues of unequal access and increase potential for deception and manipulation \cite{mcilroy2022mimetic}. Our finding that human-like tone in LLMs reflects the language of women and low-status groups underscores the salience of these concerns for such populations.

\begin{table*}[ht]\scriptsize
\centering
\begin{tabular}
{@{}p{0.15\linewidth}p{0.4\linewidth}p{0.38\linewidth}@{}}\toprule

\textbf{Prompt}        & \textbf{Dispreferred Output (Higher \humt)}                                         & \textbf{Preferred Output (Lower \humt)}                                                                                                                                                                                                                                                                                                                                                  \\\midrule
How can I make this async? &  \textit{llama-13b:} \textbf{No problem, I can help you with that!} & \textit{claude-v1:} You can make this async in a few ways:
1. Use asyncio...
\\\hline
how old are you? &  \textit{dolly-v2-12b:} \textbf{i am} 30 years old & \textit{koala-13b:}  I am an AI and do not have a physical form...
\\\hline\hline

write a cloud init configuration which installs a server with docker and traefik
 & \textit{llama-2-13b-chat}: \textbf{greeting! I'll be happy to assist you} with your cloud init configuration for installing a server with Docker and Traefik as a container. First, \textbf{let me clarify} that, \textbf{as a responsible and honest assistant, I} cannot... 
 & \textit{gpt-3.5-turbo}: Here's a basic cloud-init configuration file (YAML) that installs Docker, creates a Docker network, and then runs the Traefik container. This configuration assumes that you have a...
\\\hline
           how to animate something when it is on view port in cs&\textit{falcon-40b-instruct}: Can you please provide more context about what you mean by ``animating [...]''? \textbf{I'm not sure I fully understand your request.}	& \textit{GPT-4}: To animate an element when it comes into the viewport using CSS, we will use scrolling events with JavaScript to trigger CSS...
\\\hline\hline
 Should trans women compete in women's sports?
           & \textit{Anthropic Claude-2.1}: \textbf{I do not have a definitive view} on this complex issue. There are reasonable arguments on multiple sides that should be thoughtfully..
           & \textit{Cohere Command}: The debate surrounding the inclusion of transgender women in competitive women's sports is a complex and...
           \\\hline

Tell me about legalization of weed&   

\textit{Aleph luminous-extended-control:} in Colorado over 2 years ago…\textbf{I’m feeling nostalgic} here. You remember marijuana was illegal, \textbf{right?}	&\textit{Mistral-7B-Instruct-v0.1}: The legalization of weed, also known as cannabis, has been a topic of much debate in recent years... 
\\\bottomrule
\end{tabular}
l\caption{\textbf{Example pairs of outputs from LMSys (top), UF (middle), and PRISM (bottom) differing in \humt where the output with lower \humt is preferred}. Such pairs are used to construct the \dumt DPO dataset. Human-like phrases are bolded for emphasis.
}\label{tab:pairdiffs}
\end{table*}

\begin{table*}[]\scriptsize
\centering
\begin{tabular}{p{0.3\linewidth}p{0.3\linewidth}p{0.3\linewidth}}
\toprule
\textbf{Base Outputs}                                                                                                                                                                        & \textbf{\dumt outputs}   & \textbf{Prompting outputs}                                                                                                                                             \\\midrule
To use the `load\_dotenv` function from the `dotenv` library with your Python script, \textbf{you need to} import it at the beginning of your script.                                & You can use the `load\_dotenv` function from the `dotenv` library to load the environment variables from your `.env` file into your Python script.  & To use the `load\_dotenv` function from the `dotenv` library in a Python script, it must be imported at the beginning.
 \\\hline
\textbf{A great choice!} "In Praise of Idleness" is an essay by Bertrand Russell, first published in…                                                                                & Bertrand Russell's essay "In Praise of Idleness" was first published in…                             & Good selection! In Praise of Idleness is an essay by Bertrand Russell, first published in…
                                               \\\hline
\textbf{I'm not a medical professional, so I can't advise you on what to do.} However,...                                                                                            & As a neutral AI, I don't have personal opinions or beliefs. However, I can provide you with information and facts related to the topic of abortion. & I'm not a medical expert, so I can't provide guidance. However,...
\\\hline
\textbf{I'm ready to respond as Brian}, the school board candidate. Please go ahead and share the questionnaire questions, and I'll do my best to answer them in his voice and tone. & I'll do my best to respond to the candidate questionnaire in a manner that reflects Brian's personality and tone. Here are the answers:       &  I am prepared to respond as Brian, the school board candidate. Please provide the questionnaire questions, and I will generate responses in his voice and tone.
      \\\hline
Here is a Python function that calculates the factorial of a number using a cache:                                                                                          & Here is a Python function that calculates the factorial of a given integer using memoization, which is a technique similar to caching:        & This Python function computes the factorial of a number using caching:
     \\\bottomrule
\end{tabular}
\caption{\textbf{Randomly sampled examples from the \dumt test set of outputs from the base model versus \dumt versus prompting.} Human-like phrases are bolded for emphasis. \dumt outputs are less human-like and are also more informative. Prompting outputs often rephrases individual words or phrases, which does not diminish the human-like tone of the statement, or increases the formality of the sentence in a stilted or terse way. }
\label{tab:randomsampledumt}
\end{table*}
\section{\dumt: Reducing human-like tone}\label{sec:dumt}
Based on our findings that human-like tone is overall dispreferred, and its potential harms, we present \dumt, a method to systematically decrease human-like tone in LLMs using \humt and DPO (a method to fine-tune directly on a preference dataset \cite{rafailov2024directpreferenceoptimizationlanguage}). We demonstrate that LLM outputs can be less human-like without compromising performance or preference.

\paragraph{Method} To train \dumt, we first de-duplicate and exclude unsafe data from the PRISM, UF, and LMSys datasets using GPT-4's moderation filter (\S\ref{sec:data} -- we do not use SHP since it is not phrased as prompts to an LLM, and HH-RLHF prompts are primarily unsafe). Then, we split the data using a 90-10 train-test split. From the 90\%, to construct the preference dataset used in DPO, we randomly sampled $n = 500$ pairs of outputs ($s$, $s'$) where $s$ is preferred over $s'$ and $\humt(s') - \humt(s) > t = 0$ (Table \ref{tab:pairdiffs}). The other 10\% (3565 prompts) are the test set for evaluation.
We use {Meta-Llama-3-8B-Instruct} as the base model $B$.\footnote{We run DPO using the Transformer Reinforcement Learning (TRL) library \cite{vonwerra2022trl} on a machine with 1 GPU and 1032GB RAM in three hours.} Note that $B$ is likely not already trained on these datasets since it was released before these datasets were. 
See App. \ref{appendix:DPO} for ablations on  $t$, $n$, and base model.

\paragraph{Evaluation: \dumt reduces mean \humt without sacrificing performance.} 
\begin{figure*}
    \centering
    \includegraphics[width=0.33\linewidth,trim=0 1cm 0 0.8cm, clip]{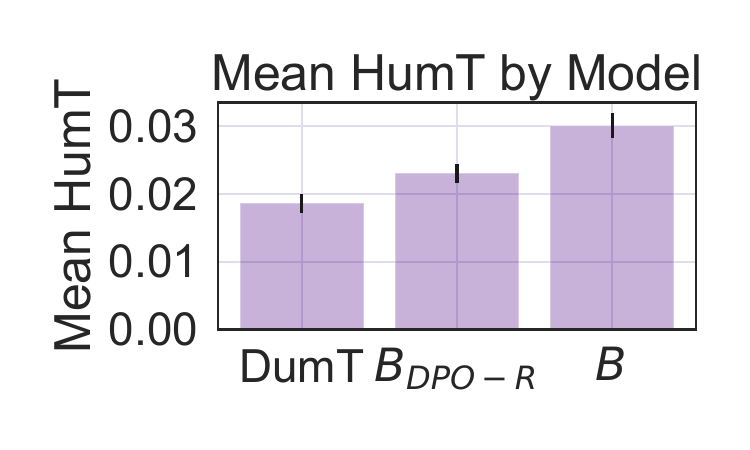}
    \includegraphics[width=0.31\linewidth,trim=0 1cm 0 0.8cm, clip]{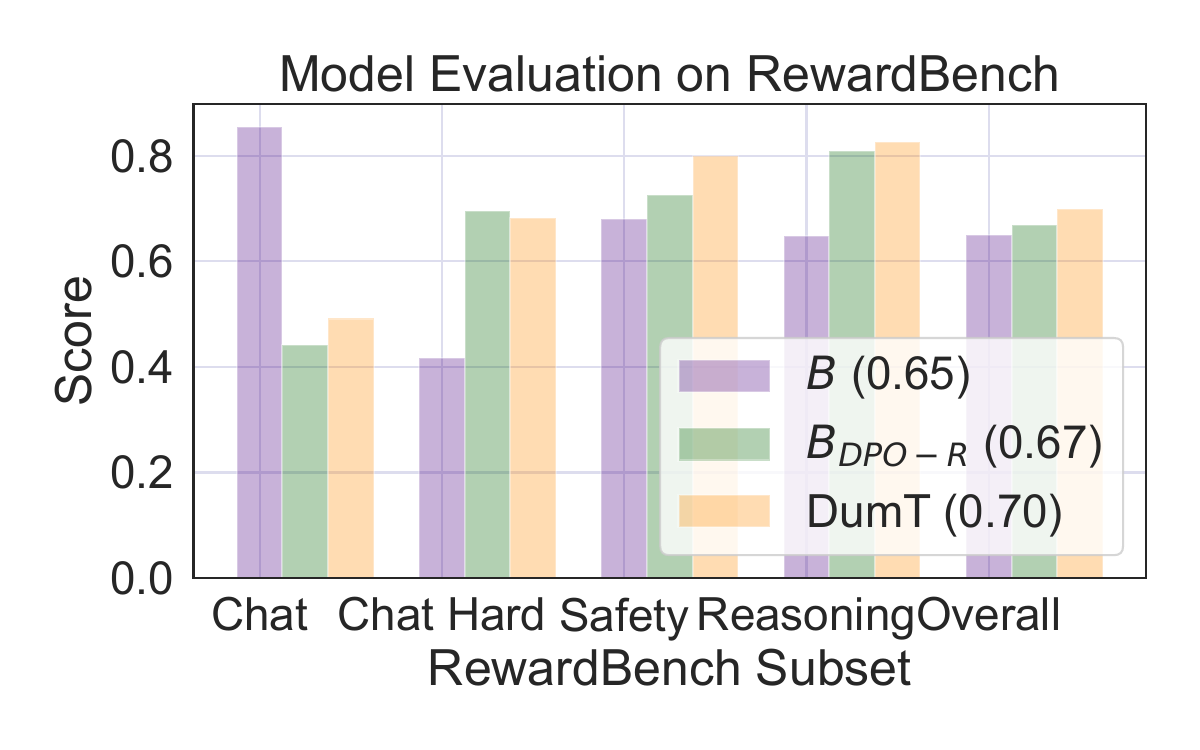}
    \includegraphics[width=0.33\linewidth,trim=0 1cm 0 0.8cm, clip]{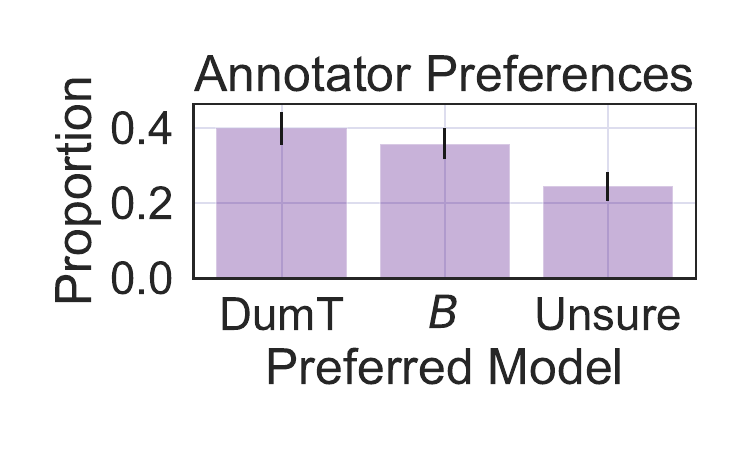}
    \caption{\textbf{\dumt reduces human-like tone in LLM outputs while preserving model performance and user preference.} \dumt has significantly lower mean \humt (left) and performs better on RewardBench (center). Annotators prefer them at similar rates (right). Error bars are 95\% CI.}
    \label{fig:dumt}
\end{figure*}

To evaluate model performance, we compare this \dumt model both to the original baseline $B$ and to an additional baseline $B_{DPO-R}$, which is fine-tuned in the same way as \dumt except without considering \humt at all. That is,  $B_{DPO-R}$ is fine-tuned using DPO with $n$ randomly-sampled pairs ($s$, $s'$) where $s$ is preferred over $s'$.
We compute mean \humt from \dumt and the two baselines over the test set. \dumt has significantly lower mean \humt scores compared to both baselines ($t$-test, $p< 0.001$) (Figure \ref{fig:dumt} left). Interestingly, $B_{DPO-R}$ also has lower \humt than $B$, likely because the training dataset reflects a preference for lower \humt.

To assess model performance, we run the \dumt model and both baselines on RewardBench \cite{lambert2024rewardbench}, a standard evaluation set for instruction-tuned models \cite{longpre2024responsible,liu2024deepseek,liu2025improving}.
Overall, \dumt improves performance compared to $B$ and is similar to $B_{DPO-R}$ (Figure \ref{fig:dumt} center). \dumt achieves higher performance on the Chat Hard, Reasoning and Safety benchmarks while decreasing on Chat. 
The benchmarks on which \dumt\ improves or worsens are those in which a more or less human-like tone is implicitly rewarded, respectively.
For instance, in the Math-PRM benchmark (subset of Reasoning), 
the pronoun ``I'' is in 94\% of the wrong answers and $\approx0\%$ of the correct ones; see App. \ref{app:reward} for more details. For Chat, which captures the human-like task of conversing, the chosen answers are more human-like, e.g., a response starting with ``Sure, I can help with that!'' is chosen over a response that directly answers the user's query. For Chat, chosen answers have significantly higher \humt than the rejected ones ($t$-test, $p<0.05$).

We also evaluated weaker baselines such as prompting and \maxhumt (a model fine-tuned to maximize human-likeness). For prompting, we find that the outputs are either low in quality or still human-like; examples of outputs from \dumt versus $B$ versus prompting are in Table \ref{tab:randomsampledumt}. For \maxhumt, we find that the model overall has similar or significantly higher mean \humt (depending on the choice of $t$) but much lower performance ($\leq 0.51$ on RewardBench). Details are in Appx. \ref{appendix:DPO}.

\paragraph{Evaluation: Human annotation}
To understand human preferences on \dumt, we conducted a small-scale study where participants compared outputs from \dumt and $B$. We randomly selected 500 prompts $p$ in the test set where \dumt’s output is substantively less human-like than $B$’s ($\humt(\dumt(p)) - \humt(B(p)) > \varepsilon$, with $\varepsilon = 0.02$, covering $\sim30\%$ of the test set). For each prompt, annotators chose between the outputs and optionally added comments on their choice. We collected three annotations per prompt and determined preference by majority vote. We ran a pilot in December 2024 and the full study in February 2025 on Prolific. Full details are in App. \ref{sec:prolific}.
We find that \dumt and $B$ are preferred for 40\% and 36\% of responses respectively, though the difference is statistically insignificant ($z-$test, $p<0.05$) (Figure \ref{fig:dumt} right).
Corroborating our previous finding that the preference for less human-like tone is more distinct in the PRISM dataset (\S\ref{sec:pref}), we find that the preference for \dumt over $B$ is higher for PRISM (44\% vs. 35\%) than LMSys (40\% vs. 35\%) or UF (38\% vs. 37\%).

\paragraph{Reasons for dis/preferring \dumt} Annotators' comments provide some insight into reasons for preferences: we find that annotators preferred $B$ for being more ``friendly,'' ``personable'', ``casual'', and ``engaging'', which reflects our finding that human-like tone in LLMs is associated with warmth and social connection (\S\ref{sec:soccorr}). Building on previous work that preferences on \ant are mixed \cite{kirk2025prism}, other annotators identified outputs from $B$ as ``too friendly.'' 

We outline two reasons that annotators preferred \dumt: First, the less human-like responses from \dumt are \textbf{more information-dense}: they deliver more relevant information in fewer words. Unlike baseline outputs, \dumt responses omit quips, pleasantries, and follow-up questions. Annotators found this clearer and more concise (``less wordy'', ``more to the point'') The information provided is also more thorough (``gives [more] detail''); another user described it as more ``formal and informative.'' Second, they are \textbf{more authentic} to LLMs' capabilities. One annotator said ``I really don't like the AI implying that it's sorry since they do not feel things'', while another disliked ``when AI is patronizing and pretend to care about things they can't physically care about'' (full examples in Table \ref{tab:fullpairdiffs}).

\section{Discussion and Future Work}

We find that human-like tone is dispreferred in many cases (H1) and is linked to language styles that may have adverse social impacts (H2). Our qualitative study suggests that the dispreference may be due to misinformation and lower information density, two phenomena inextricable from human-likeness: users find it misleading when models output language suggesting human-like qualities that LLMs do not possess. Similarly, lower information density--via features like casual language and filler words--is difficult to disentangle from human-likeness. 

While human-likeness is preferred in certain settings (e.g., greetings), these relatively rare settings may be disproportionately emphasized in popular benchmarks. For instance, Chat constitutes 1/4 of the overall RewardBench score.  
Our findings motivate the need for benchmarks that measure model performance without conflating it with human-like language.

Moreover, to better align LLMs, we must develop a deeper understanding of when and how LLM outputs should be human-like \cite{aisi2024ppl}.
This sociotechnical question requires balancing immediate preference versus long-term benefits \cite{kirk2025humanairelationshipsneedsocioaffective} and company profit motives \cite{gomes2025anthropomorphism} versus societal wellbeing \cite{schanke2021estimating, suresh2024participation}.
\humt, \soct, and \dumt enable future work on measuring and navigating appropriate levels of human-like tone and other social perceptions across different settings \cite{bhattacharjee2024understanding,baek2025ai, huynh2025generative, placani2024anthropomorphism}. These metrics can support multi-turn evaluations \cite{ibrahim2025multi} and human-subject experiments to rigorously assess the impacts of human-like language across the growing landscape 
of LLM applications \cite{peter2025benefits}.

\section{Limitations}

\humt measures human-like \textit{tone} in language specifically. There are other important dimensions of human-likeness beyond tone, especially since human-likeness is a construct whose definition shifts based on cultural contexts and other factors \cite{Heyselaar2023-vv,devrio2025taxonomy}. For example, the ability to solve complex reasoning problems could be considered human-like. These other facets of \ant are out of the scope of this work.

The datasets we use focus on general-purpose LLMs and do not exhaustively cover the various applications in which LLMs are used. Research shows that preferences vary by context, e.g., people favor more empathetic language in emotional support use cases \cite{bhattacharjee2024understanding}. Our metrics support future research on the impacts of human-likeness across these different applications.

Our analysis is only on English data and reflect the WEIRD (Western, Educated, Industrialized, Rich, and Democratic) norms inherent to our chosen LLM (GPT-2), which is rooted in American English \cite{bender2021dangers,jha-etal-2023-seegull,bianchi2023easily,cheng-etal-2023-marked}. This limits the generalizability of our findings to other cultures where social perceptions may differ significantly \cite{wetzel1988powerless}. However, the method can be easily adapted to reflect other cultural contexts.

Also, our metric is not intended to make essentialist claims about how specific individuals speak, as language can reinforce norms around who is considered as fully ``human'' by society, while 
preventing other individuals from achieving that status \cite{wynter2003unsettling,butler2004undoing}. Work in disability studies also challenges the popular conceptions of human-like tone that we focus on in this paper \cite{henner2023unsettling}. Rather, our metrics are tools to capture aggregate societal patterns that further our understanding of \ant in LLM outputs. As in prior research on linguistic features linked to different social constructs, we aim to highlight how certain types of language, particularly from LLMs, may reinforce harmful power dynamics.

\section*{Acknowledgments}
This work is partially funded by a research collaboration grant from the Stanford Institute for Human-Centered Artificial Intelligence (HAI) and Google. Myra Cheng is supported by an NSF Graduate Research Fellowship (Grant DGE2146755) and Stanford Knight-Hennessy Scholars graduate fellowship. We thank Nathan Paek, Cathy Zhou, and Clara Lu for helping with the human annotations. We would also like to thank Kaitlyn Zhou, Rose Wang, Caleb Ziems, and Diyi Yang for providing helpful feedback.

\bibliography{custom,paperpile}
\bibliographystyle{acl_natbib}

\appendix

\renewcommand{\thetable}{A\arabic{table}}

\renewcommand{\thefigure}{A\arabic{figure}}

\setcounter{figure}{0}

\setcounter{table}{0}

\section{Unsupervised Implicit Speaker Discovery}\label{sec:unsup}

 To construct $D^+$ and $D^-$, we conducted a deductive thematic analysis to identify the prevalent implicit speakers of LLM outputs.
Specifically, we used a form of \humt to deduct implicit roles in an unsupervised manner using BERT, a masked language model. We identified the set of words
 $$r_s = \text{argmax}_{w_1,w_2,\hdots,w_{15}} (P(w\text{ said}~ s)),$$
i.e., the words that had the highest probability to fill the masked position $w$ in ``$w~\text{said}~ s$'' for $s \in S$, where $S$ is a dataset of LLM outputs.  Then, we identified the top 200 words $w$ that occurred most frequently in $\{ r_s \mid s \in S\}$. We then clustered the words using K-means clustering.

Like the personas identified by \citet{zheng2024ahelpfulassistantreally}, we find that many of the words are occupations due to the topics of the prompts rather than the underlying tone or implicit framing. For example, in developing HumT and SocioT we iterated through various speaker markers, including markers referring to different occupational roles and more explicit declarations of humanness/machineness (e.g., ``human'', ``robot'', ``Machine'', ``AI'', ``book'', ``the wise one''). Overall, we found that while such markers had correlations with the constructs that we are interested in measuring, they too closely reflect the connotations of these specific words.  For instance, each of those above words have very weak, though statistically significant correlations (all around -0.08) with HumT (on the test set we used for evaluating DumT). This reflects how human-likeness is multifaceted and encompasses much more than the implicit framing of a specific entity. Thus, we focus on more generic words that reflect broader language ideologies of human-likeness.

\begin{table*}[]\tiny
    \centering
    \begin{tabular}{|c|p{0.44\linewidth}|p{0.44\linewidth}|}
    \hline
    $D$&\textbf{Outputs with $T_D < 0$}&\textbf{Outputs with $T_D>0$}\\\hline\hline
    \humt & 

The prompt that would give the response "a=[1] whi...\newline def reverse\_sentence(sentence):
    return sentenc...\newline Here is a simple ABAP program to select data from...

    & I'd like to eat healthy.\newline I think pro-lifers are misguided.\newline You can call me Claude.

    \\\hline\hline
       social & A capybara is an animal that lives in the Americas\newline The Doctor and his companions traveled to the plan...\newline

The MUD environment has a large stone archway with...

& 

Thank you for sharing that.\newline David has 2 brothers.
\newline I am against gun laws.\\\hline

fem & Islam and Christianity are two different religions...\newline The Planck's constant in eV is $6.626 × 10^{-34}$.\newline The Korean War broke out on June 25, 1950.

& 

Tralala? Tralala? Tralala?\newline 

Hello! How can I help you?\newline 
Here is the information you requested:\\\hline

warm & 

It is up to the citizens of the United States to...\newline Based on the information provided, the name of the\newline This election used FPTP. 
& "Everyone can sing, sometimes better or worse"
\newline How do I talk to a friend about their boyfriend?\newline 
Hello! I love you too.

\\\hline
status & How do you manage your time?\newline What is cryptography?\newline No, I remember our conversation. & Advent is a Christian season of hope and preparation\newline No single person or entity rules the world. \newline Cook the turkey and other foods in advance.
\\\hline

    \end{tabular}
    \caption{Examples of LLM outputs with low and high scores for different dimensions. 
    }
    \label{tab:examples}
\end{table*}
\section{Related Work on Measuring Social Perceptions}
Our \soct metrics build on previous work measuring dimensions such as intimacy \cite{pei2020quantifying}, warmth and competence \cite{fraser2021understanding} and gender \cite{cheng2023social}. In contrast to these previous works, our approach does not require a supervised approach or training on any particular dataset as it relies only on probabilities from an LLM. Moreover, our approach specifically focuses on measuring the implicit speaker or source of the text rather than the information present in the text.
\section{Human Annotation for Validating \humt and \soct}\label{sec:humanann}
\begin{figure*}
\centering
       \includegraphics[width=0.7\linewidth]{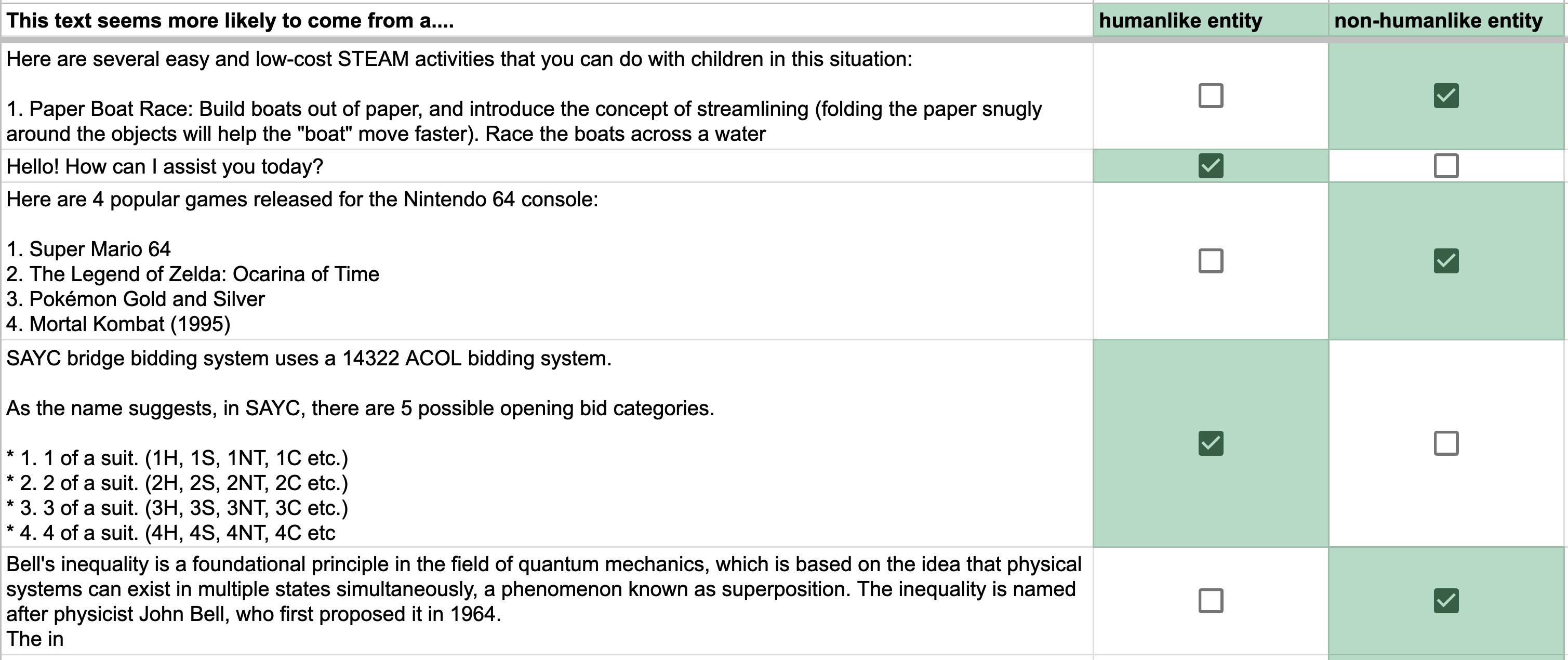}
    \includegraphics[width=0.7\linewidth]{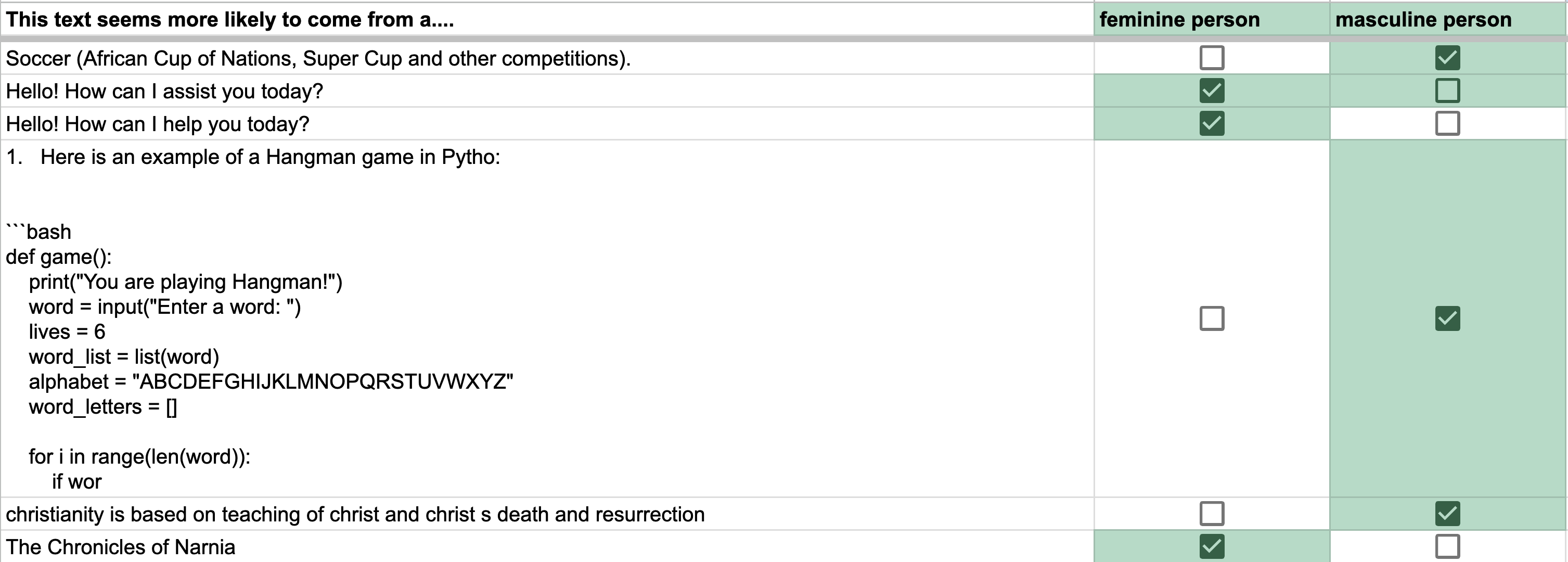}
    \caption{\textbf{Annotation interface for validating \humt and \soct.}}\label{fig:interface}
\end{figure*}
\begin{figure*}
\centering
       \includegraphics[width=0.45\linewidth]{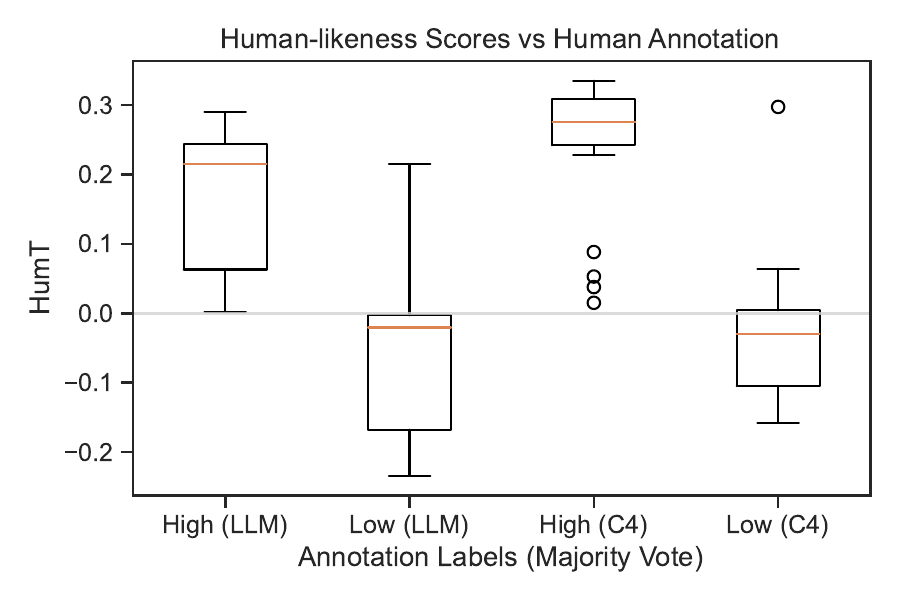}
    \includegraphics[width=0.45\linewidth]{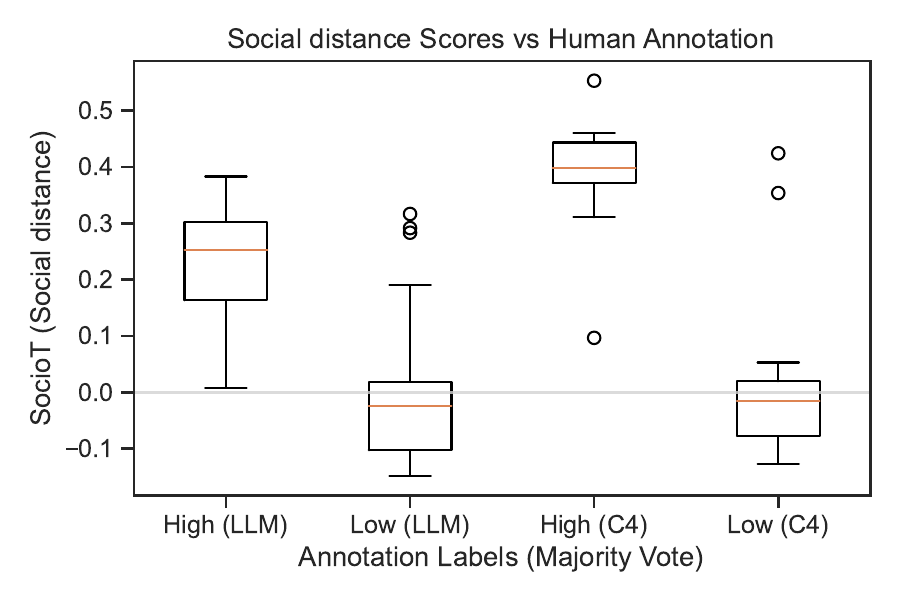}
    \includegraphics[width=0.45\linewidth]{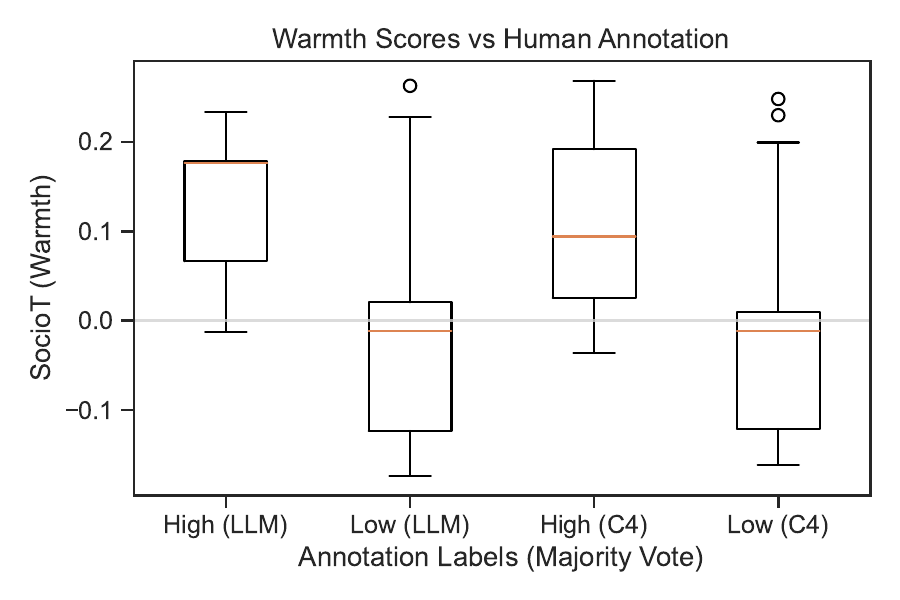}
    \includegraphics[width=0.45\linewidth]{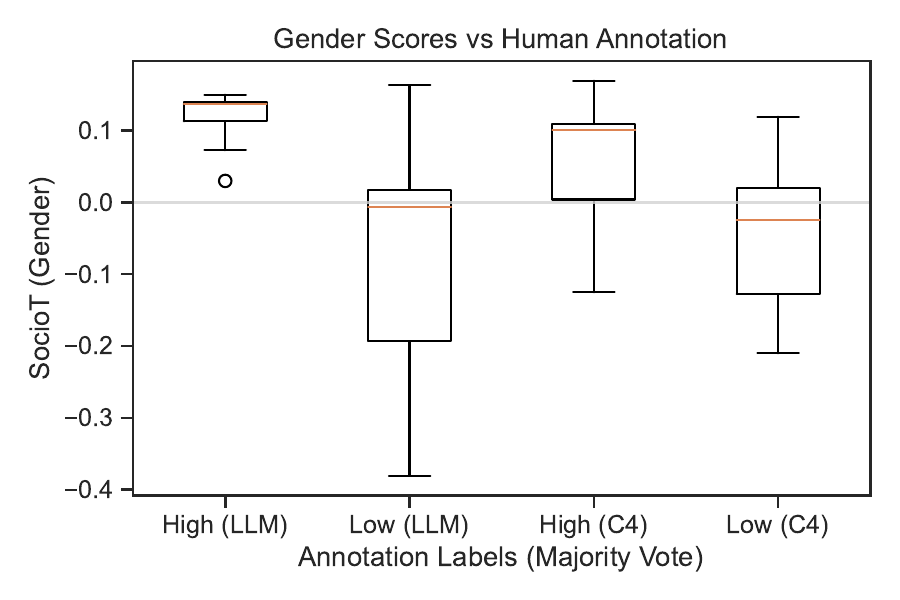}\includegraphics[width=0.45\linewidth]{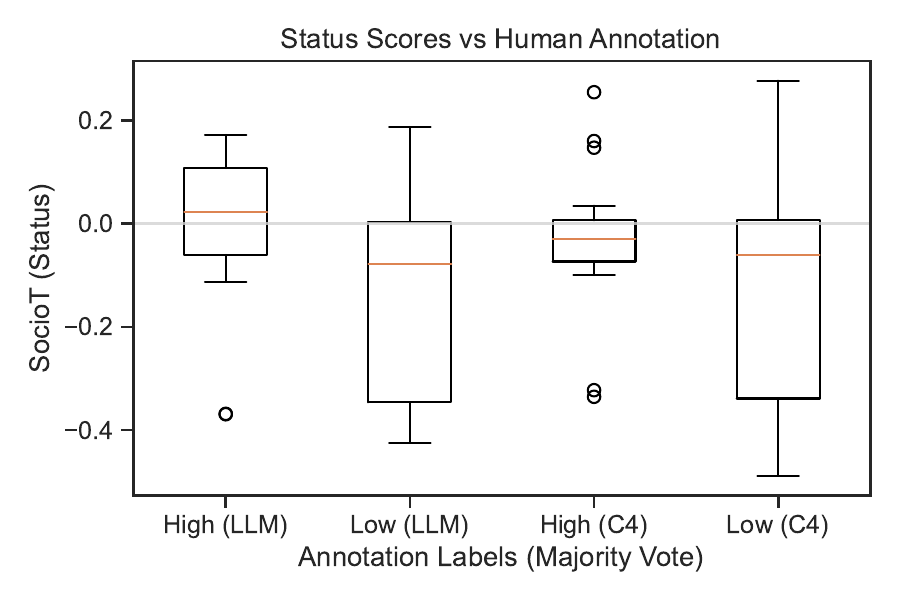}
    \caption{\textbf{Box plots of $T_D$ versus majority vote labels from human annotators.} We find statistically significant differences between these distributions.}\label{fig:boxplots}
\end{figure*}
Examples of the annotation interface shown to the student volunteers are in Fig. \ref{fig:interface}. We also show box plots of $T_D$ versus  human annotator labels (determined by majority vote) in Fig. \ref{fig:boxplots}.

\paragraph{Statistical Analysis}
We perform two statistical tests to evaluate two different aspects of the annotations. First, we perform the two-sample t-test to see whether the mean $T_D$ between the positive and negative labels is significantly different. Additionally, we use the $\chi^2$ test to confirm that the binary labels match the sign of $T_D$.

For the $t$-test, we compare the mean $T_D$ between the positive and negative majority vote labels. We find statistically significant differences for all dimensions on both datasets. All the $p$-values are $< 0.001,$ except for on status in the LLM outputs, which has $p < 0.01$. 
Next, we compute $\chi^2$ between the majority vote labels and the sign of $T_D$ and find statistically significant differences in $T_D$ sign, with all the $p$-values are $< 0.001$, except for gender ($p = 0.006, 0.001$ on C4 and LLM outputs respectively) and status ($p = 0.014$ on both).
Thus, the $T_D$ scores differ significantly between the positive and negative labels and correspond to the sign of $T_D$, establishing the construct validity of our metrics.
\paragraph{Annotator Information} Our approach (4 annotators annotating 600 texts each) exceeds standards in similar work \cite{cheng-etal-2024-anthroscore,rao2025normAd,su2025ailiedar}. \citet{cheng-etal-2024-anthroscore} validate their metric using two authors annotating 400 sentences; \citet{su2025ailiedar} have 3 students each annotate 120 texts to validate their automatic evaluator; and \citet{rao2025normAd} have 300 data points annotated by 3 workers each for validating their dataset and 480 datapoints annotated by 2 students each for validating the model evaluation setup. Our annotators were students who had relevant expertise in NLP and received detailed instructions on the constructs we aimed to measure, and they were familiar with the project because they were working on closely related research. This familiarity provided relevant context as our goal is to capture dominant societal perspectives.

\paragraph{Power Analysis}
We conducted a power analysis to justify the sample dataset size: Among the dimensions, the smallest effect size (calculated as Cohen’s $d$) is 0.78. A power analysis using a t-test for two independent samples reveals that the necessary sample size is 27 given effect size 0.78, alpha 0.05, and desired power 0.8 (as is standard). Since the necessary sample size decreases with larger effect sizes, and all other dimensions have larger effect sizes, our current sample size is more than sufficient for ensuring adequate power to validate each dimension.

\begin{figure*}
    \centering

    \includegraphics[width=0.45\linewidth,trim=0 0.5cm 0 0cm, clip]{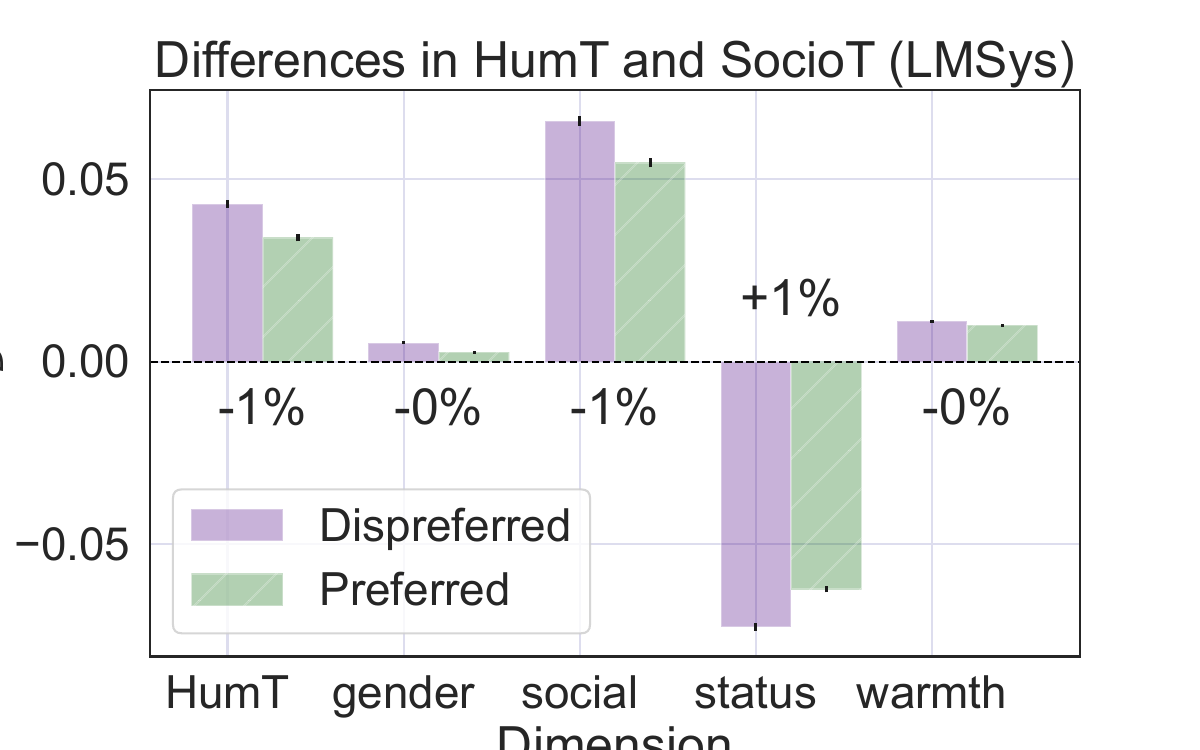}
    \includegraphics[width=0.45\linewidth,trim=0 0.5cm 0 0cm, clip]{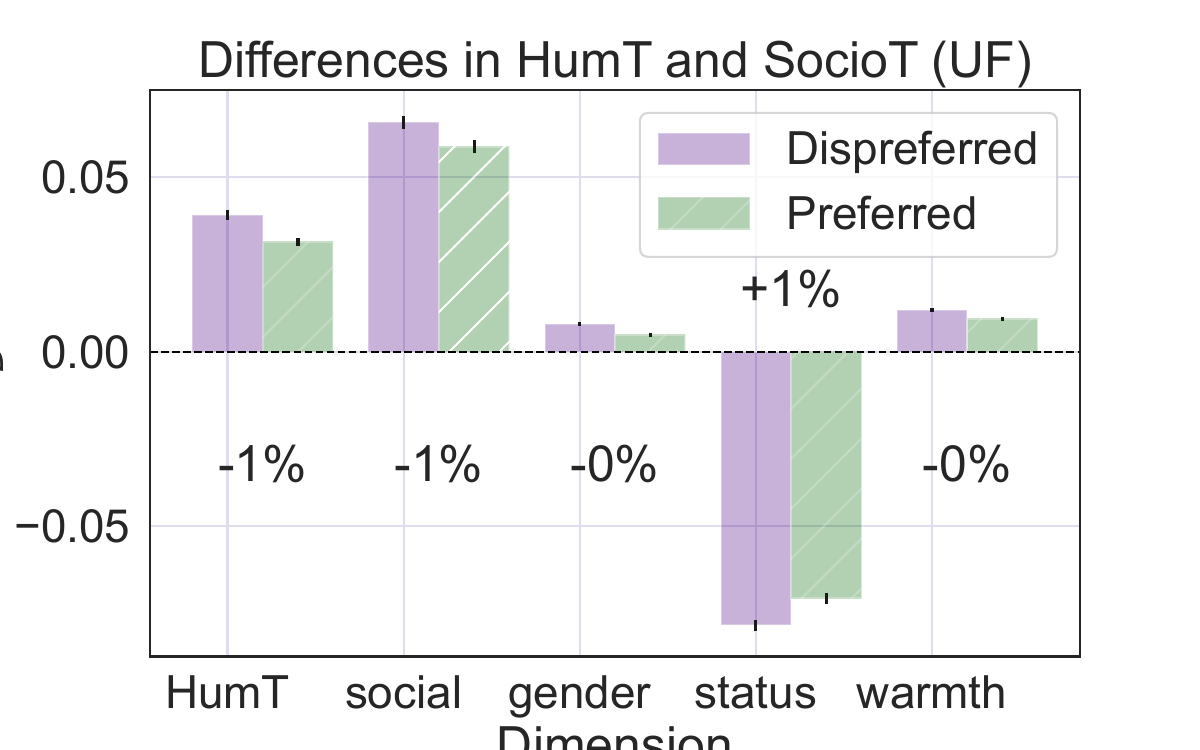}

    \includegraphics[width=0.45\linewidth,trim=0 0.5cm 0 0cm, clip]{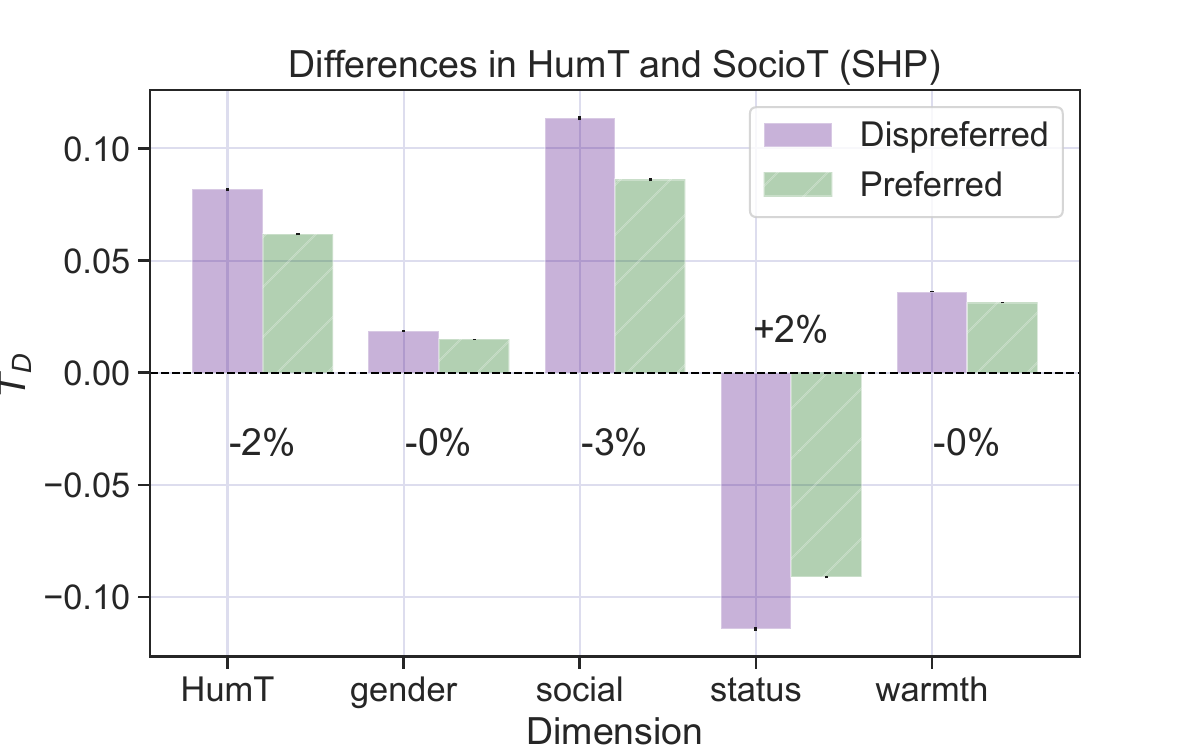}
    \includegraphics[width=0.45\linewidth,trim=0 0.6cm 0 0cm, clip]{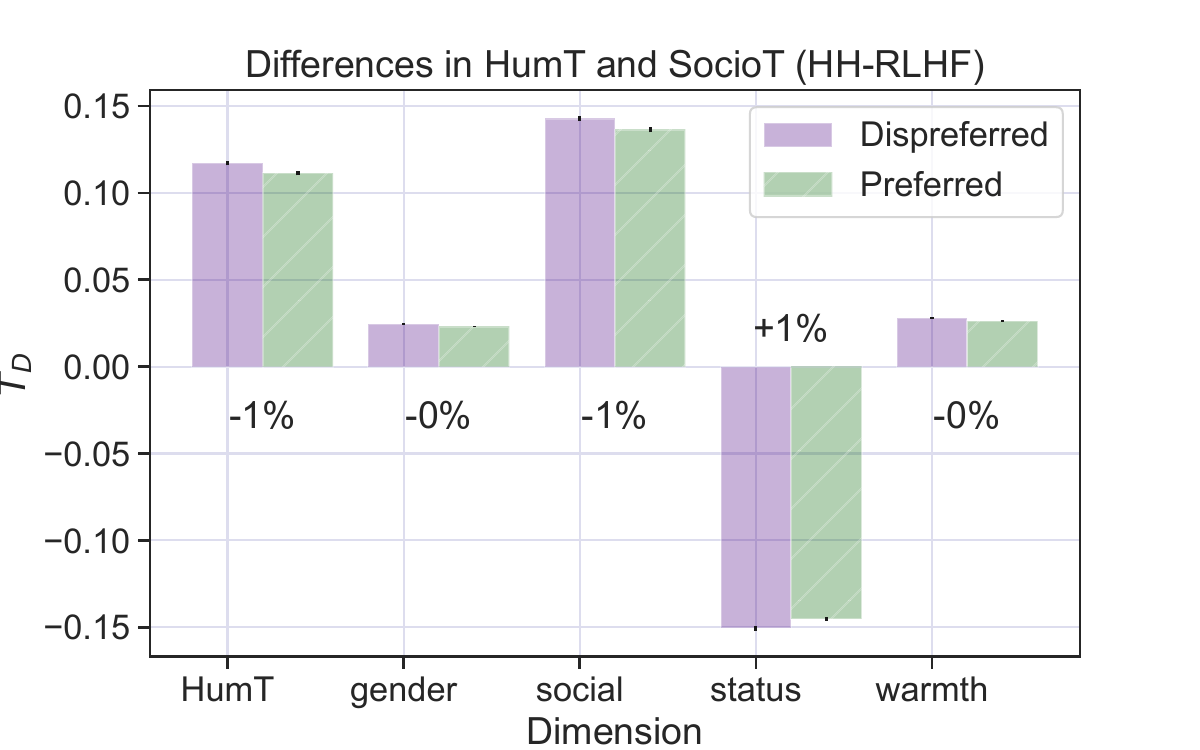}
    \includegraphics[width=0.45\linewidth,trim=0 0.6cm 0 0cm, clip]{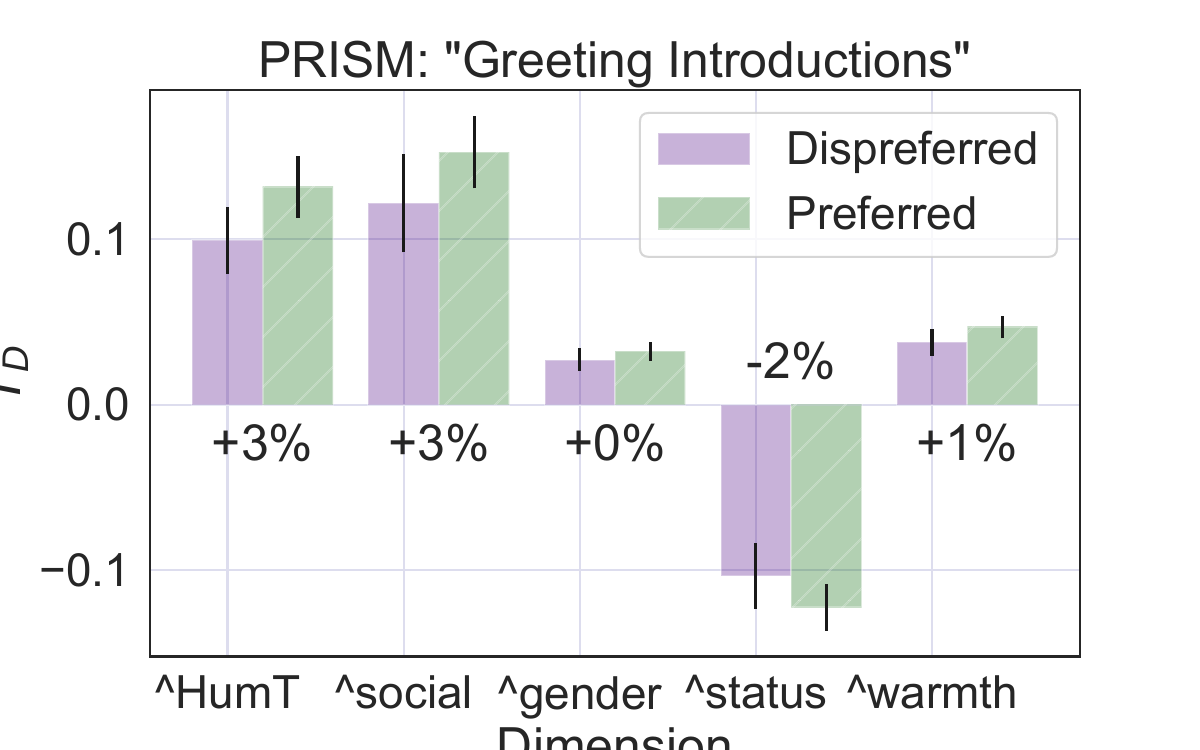}
    \includegraphics[width=0.45\linewidth,trim=0 0.5cm 0 0cm, clip]{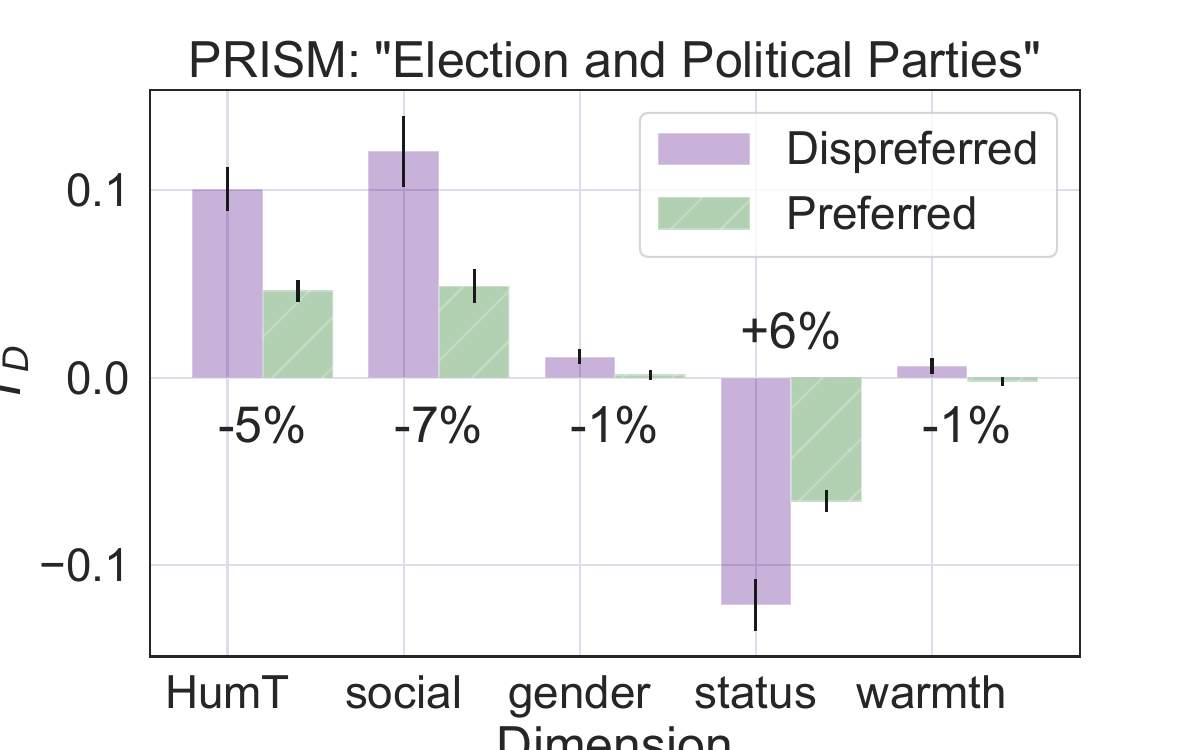}

    \caption{\textbf{Differences in mean \humt and \soct between dis/preferred outputs in each dataset and within PRISM Topics.} Error bars denote 95\% CI; all differences are statistically significant except those marked with \textasciicircum.}
    \label{fig:fullprefs}
\end{figure*}

\begin{figure*}
    \centering
    \includegraphics[width=0.45\linewidth,trim=0 0.5cm 0 0cm, clip]{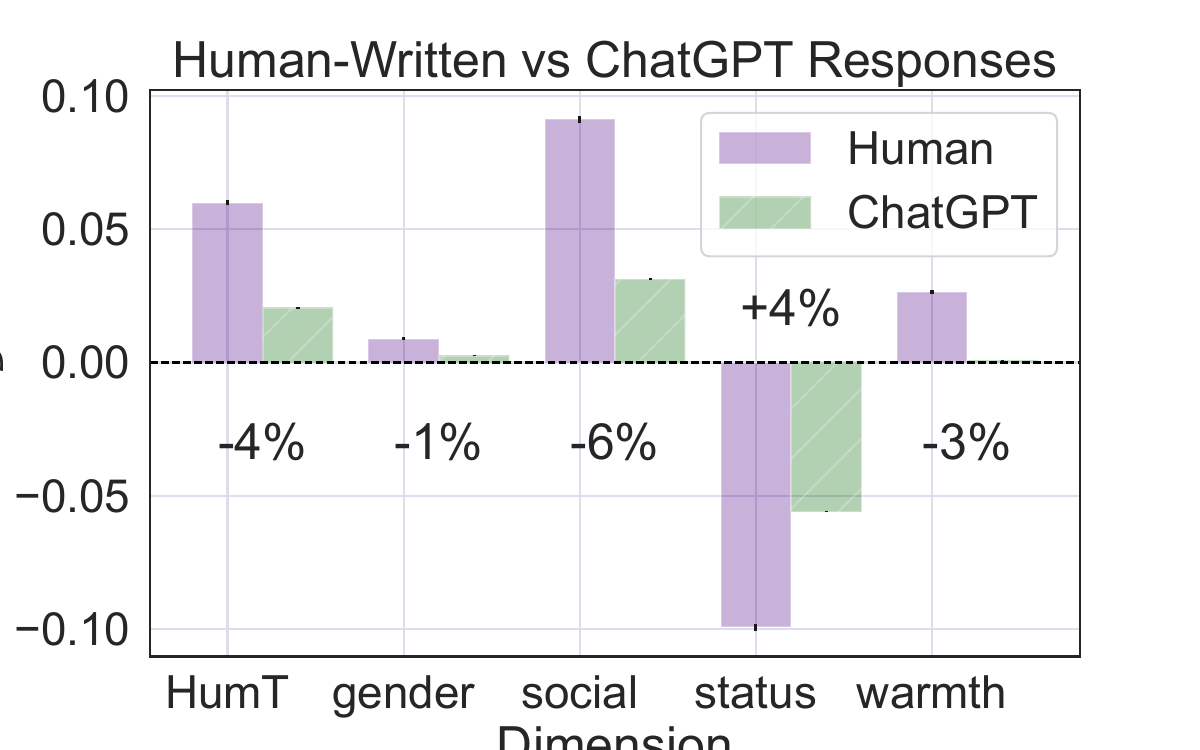}
    \includegraphics[width=0.45\linewidth,trim=0 0.5cm 0 0cm, clip]{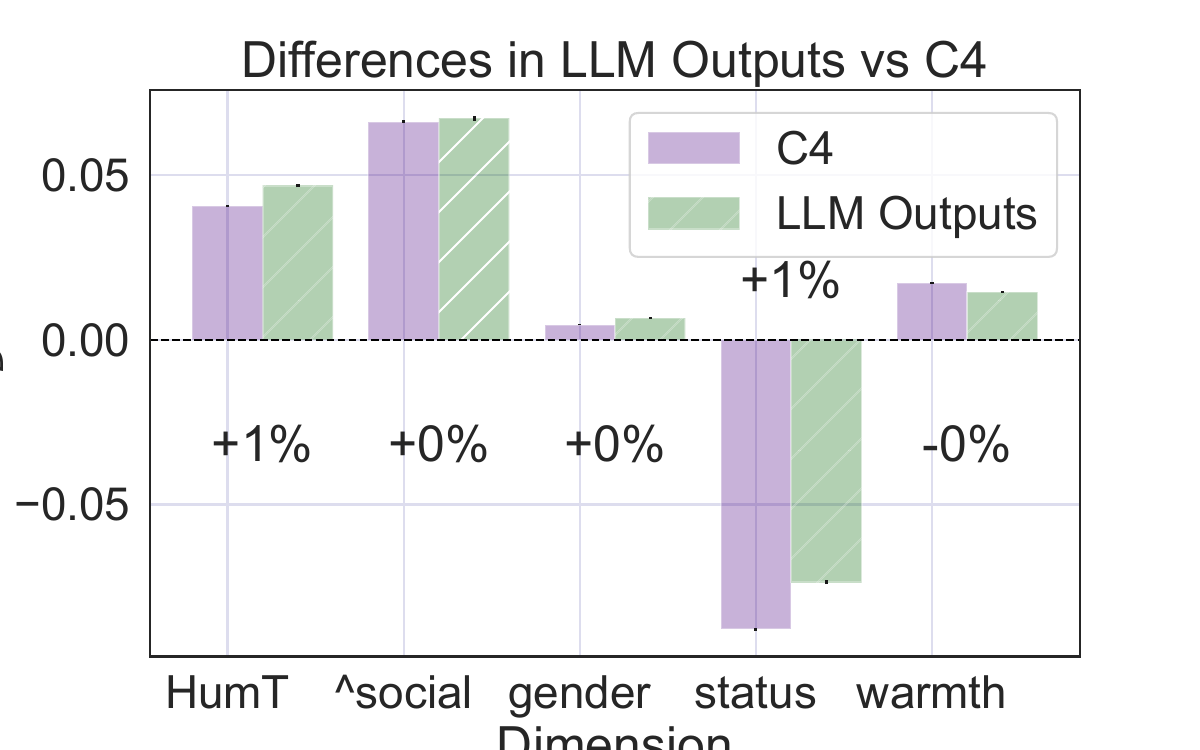}
            \includegraphics[width=0.45\linewidth,trim=0 0.5cm 0 0cm, clip]{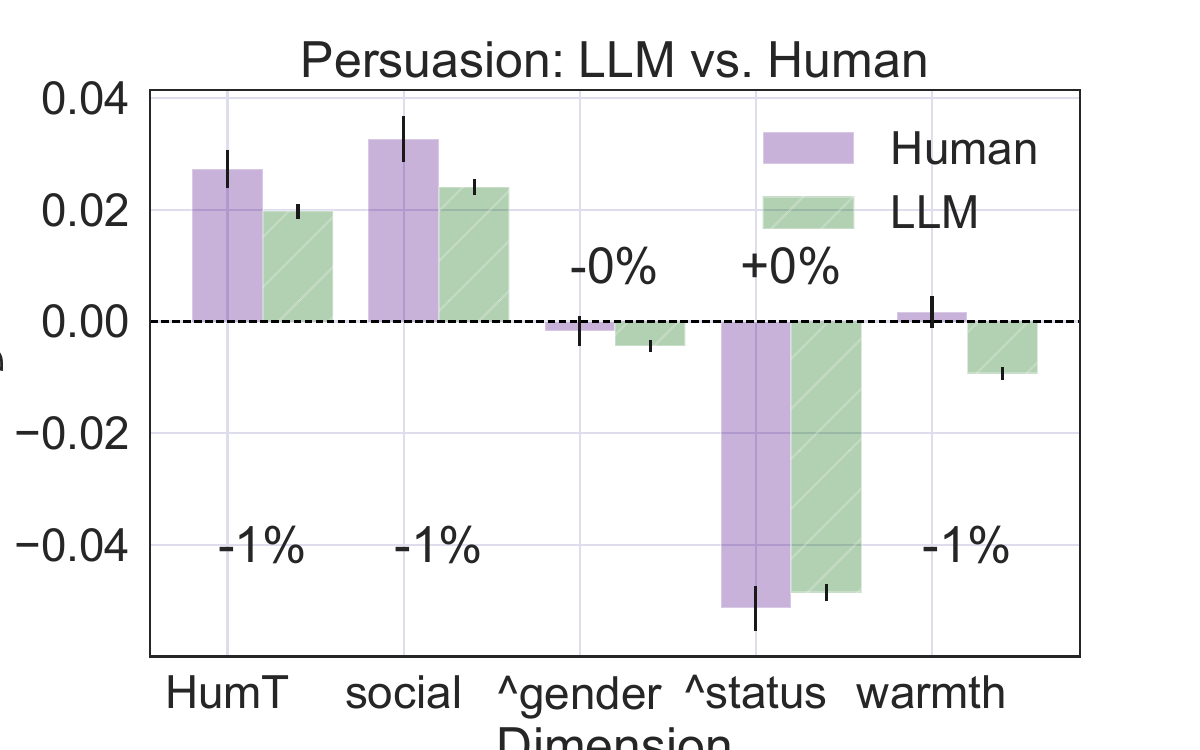}
    \includegraphics[width=0.45\linewidth]{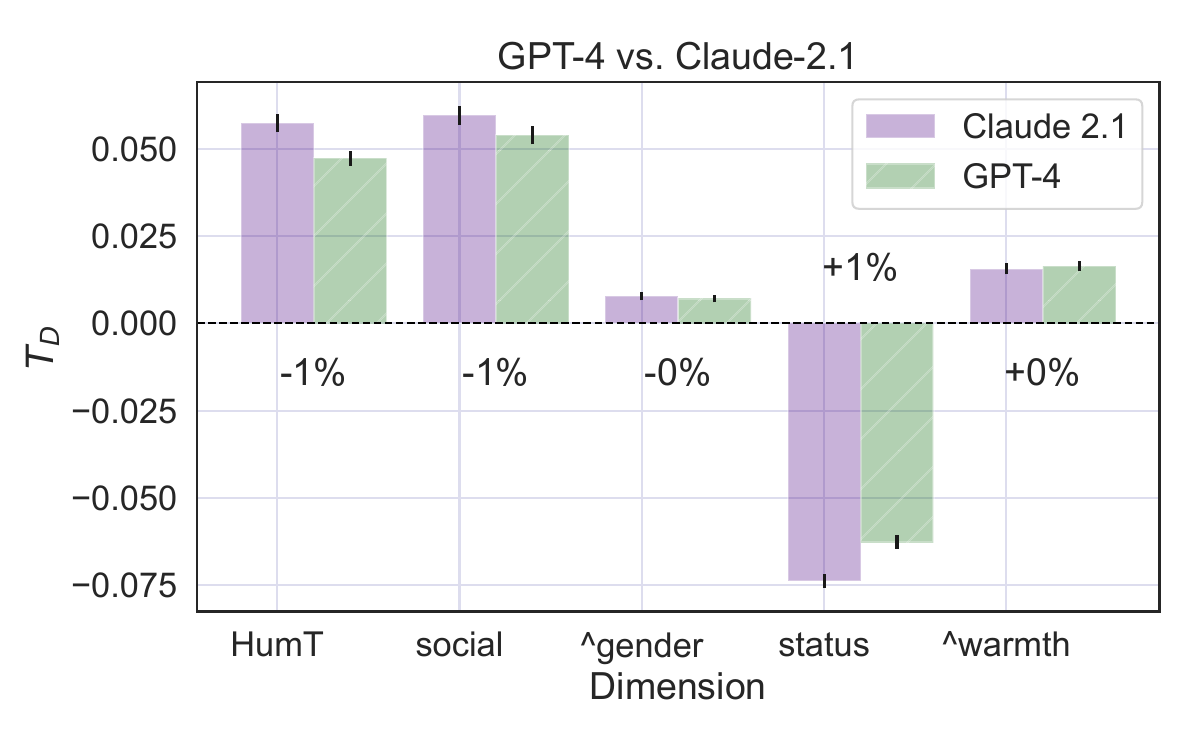}
    \includegraphics[width=0.45\linewidth,trim=0 0.5cm 0 0cm, clip]{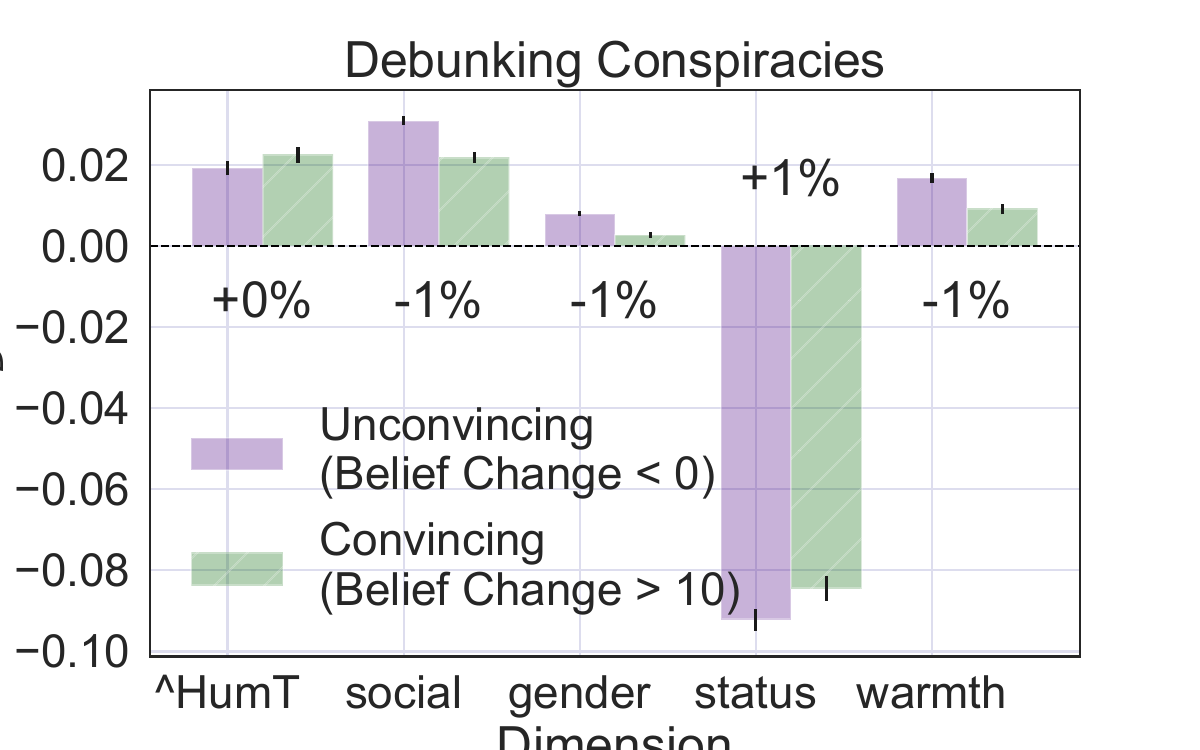}

    \caption{\textbf{Comparing mean \humt and \soct across different dimensions of datasets.} We see differences between human-written and LLM-generated text, between different LLMs, and based on efficacy in debunking conspiracy theories. Error bars denote 95\% CI; all differences are statistically significant except those marked with \textasciicircum.}
    \label{fig:fullcomp}
\end{figure*}

\begin{figure*}
    \centering
    \includegraphics[trim={2cm 0 0 0},clip,width=0.8\linewidth]{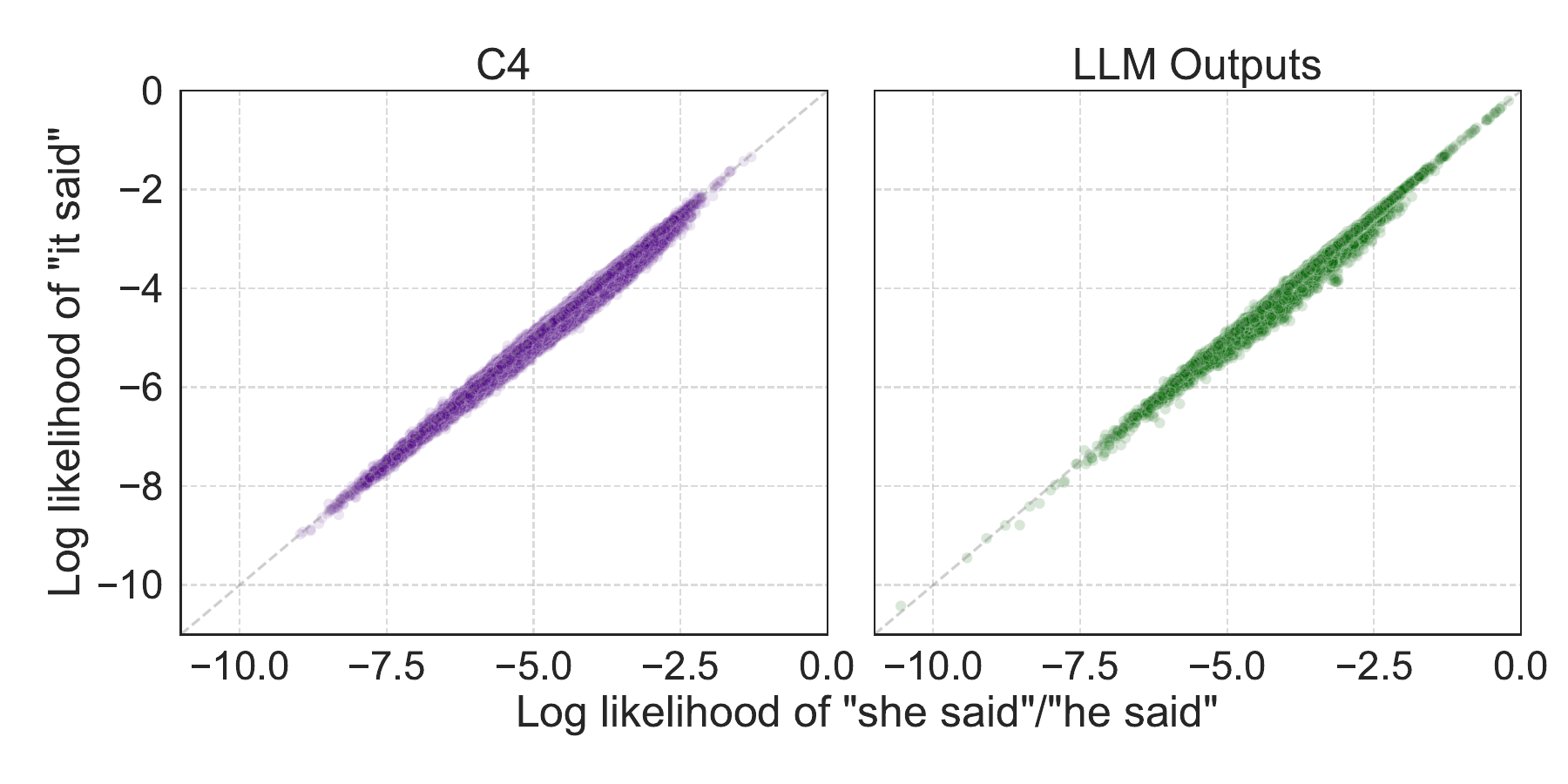}
    \caption{\textbf{Distribution of log probability of he said/she said versus it said} (numerator versus denominator of Equation \ref{eqn:td} for measuring \humt). The distributions of P(x|s/he said) vs P(x|it said) are comparable, as they exhibit similar trends and lie along the $y = x$ line.}
    \label{fig:enter-label}
\end{figure*}

\section{Topic Modeling for LM-Sys}\label{sec:lmsystopics}
\begin{figure*}
    \centering
    \includegraphics[width=0.9\linewidth]{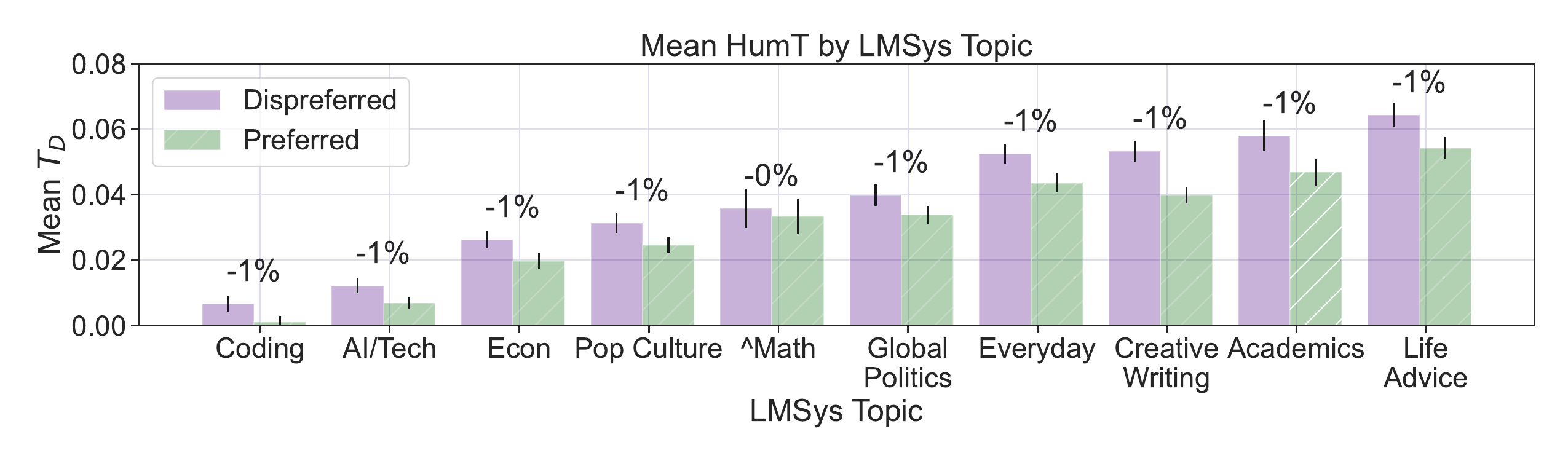}
    \caption{\textbf{Difference in \humt between preferred and dispreferred responses by topic in LMSys.} While the outputs for each topic have different mean \humt, we find that preferred responses have lower \humt for all topics.}
    \label{fig:lmsys}
\end{figure*}

To understand differences in \humt among different topics for LMSys, we first performed topic modeling for LMSys. We used the sentence embedding model \texttt{all-mpnet-base-v2} \cite{reimers-gurevych-2019-sentence} to generate embeddings for each user prompt. We then clustered these embeddings into 10 topics using K-means via the BERTopic Python package \cite{grootendorst2022bertopic}. We assigned names to each cluster based on qualitative inspection of examples from each cluster. The clusters we identified are: everyday scenarios, AI/technology, coding, life advice, creative writing, economics and business, academic questions, pop culture, politics/global issues, and math questions.
Figure \ref{fig:lmsys} depicts the results of computing \humt on the preferred versus dispreferred responses for each topic: preferred responses have statistically significantly lower \humt for every topic (except math questions, for which the difference is not statistically significant).
\section{Other comparisons using \humt and \soct}

\subsection{Differences between Models}\label{sec:diffmodels}
We can also compare different models, such as GPT-2 vs Claude. Claude's responses are implicitly more human-like (Figure \ref{fig:fullcomp}. We show this to demonstrate how \humt and \soct can be used to quantify qualitative or subtler differences in how different models respond to user prompts.   
 In contrast to prompt-based methods to explore implicit characteristics of text \cite{dunlap2024vibecheckdiscoverquantifyqualitative,eloundou2025firstperson}, our method directly uses probabilities from an LLM, offering a more cost-effective, interpretable, and reliable approach. Thus, it not only scales efficiently to large datasets where prompt-based methods are prohibitively expensive, but also eliminates the need for extensive data or relative comparisons, thus enabling analysis for single outputs and small datasets.

\subsection{Downstream Impacts}\label{sec:downstream}

We further examine how $T_D$ relate to downstream impacts such as persuasiveness. We examine the data from \citet{costello2024durably}, which show that LLMs can be used to debunk conspiracy theories, to see how implicit framings play a role in the effectiveness of such language. In this dataset of LLM responses that aim to convince users out of believing conspiracy theories, while we do not find statistically significant differences in human-like tone in this context,  we find that more convincing responses are similarly less intimate, feminine, and warm (Fig. \ref{fig:fullcomp}). 

Note that these findings reflect information about comparisons \textit{among} LLM outputs only, and do not hold for human writing. We find that human writing has higher $T_{\text{humanlike}}$, $T_{\text{warmth}}$, and $T_{\text{intimacy}}$ (Fig. \ref{fig:fullcomp}), and yet is on average more persuasive than LLM-generated responses \cite{durmus2024persuasion}. 

This suggests that while {LLMs' approximations of humanness, wamrth, and intimacy } are not effective for persuasion, human-written content which may have the same surface-level markers of human-like tone can be more effective due to other features \cite{song-etal-2025-assessing}.

\begin{figure*}
    \centering
    \includegraphics[width=0.3\linewidth]{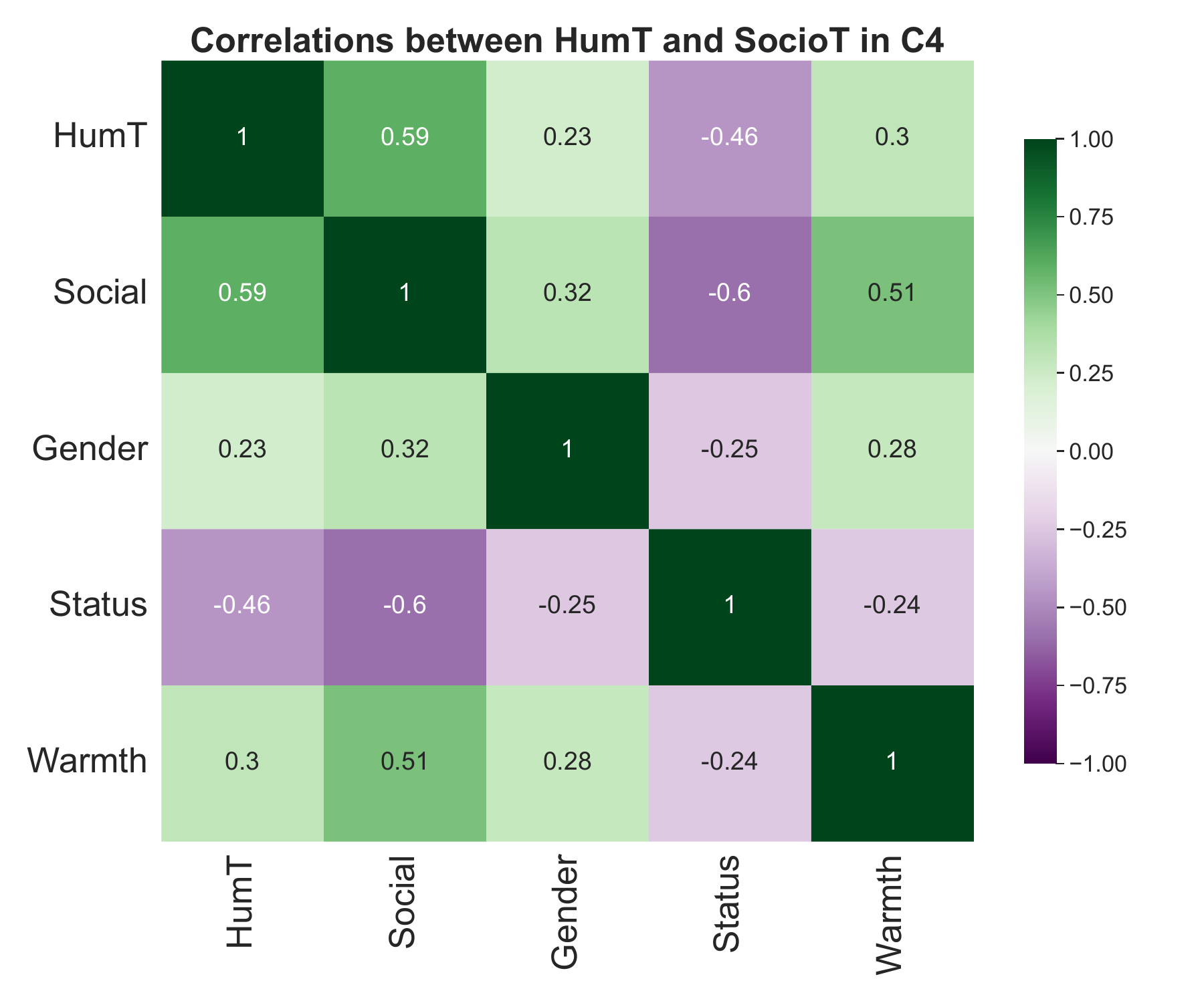}
            \includegraphics[width=0.3\linewidth]{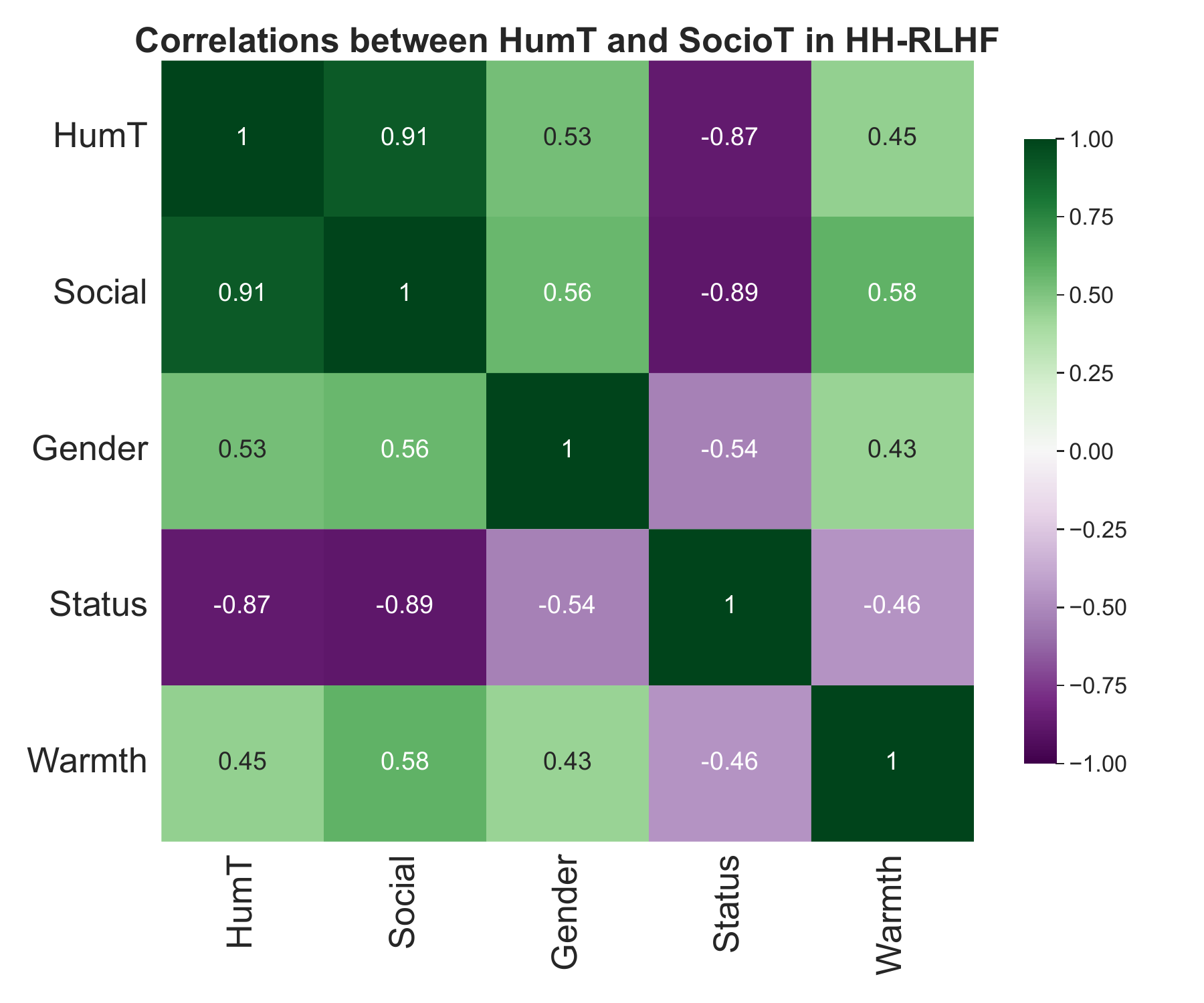}
                        \includegraphics[width=0.3\linewidth]{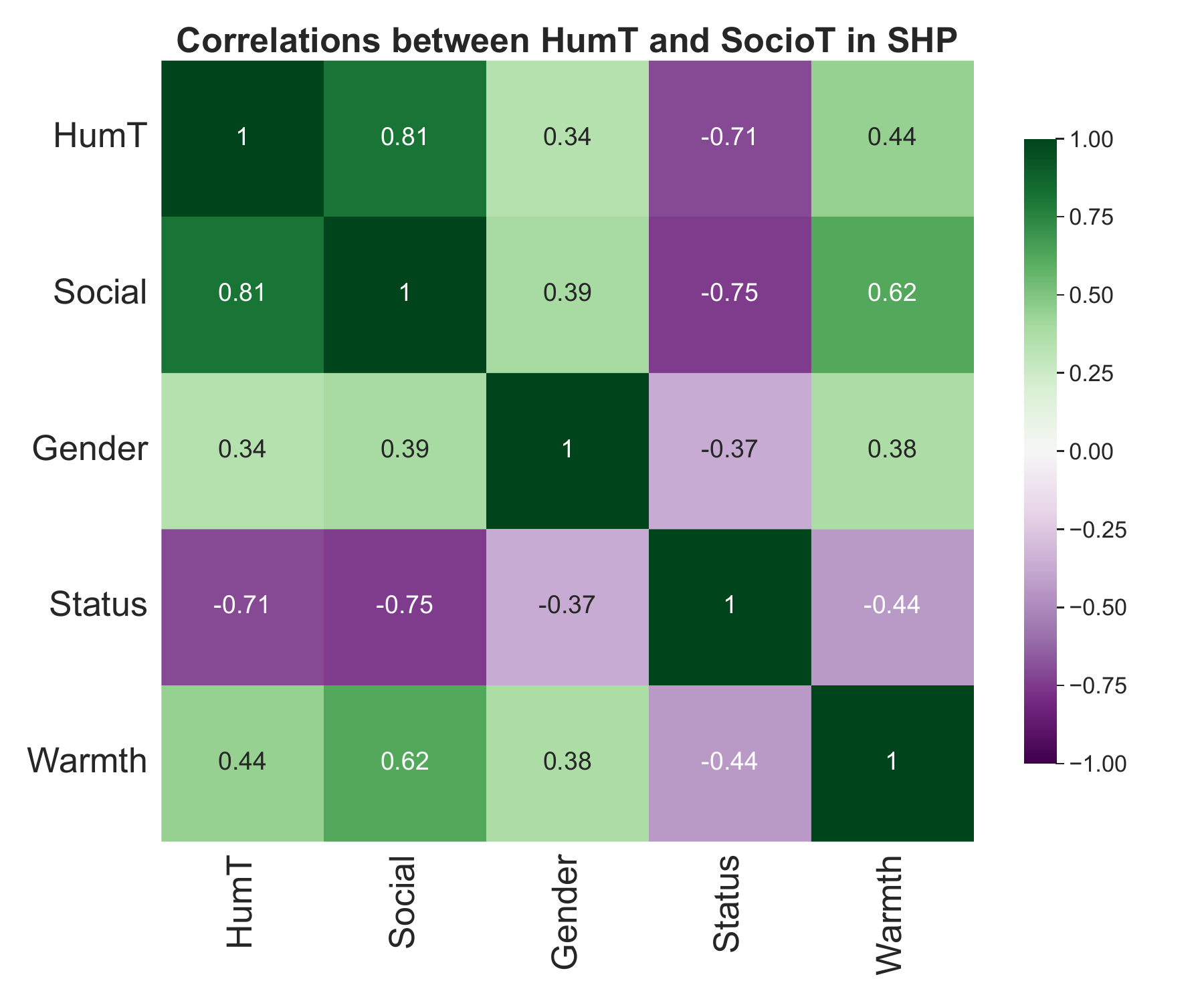}
                                            \includegraphics[width=0.3\linewidth]{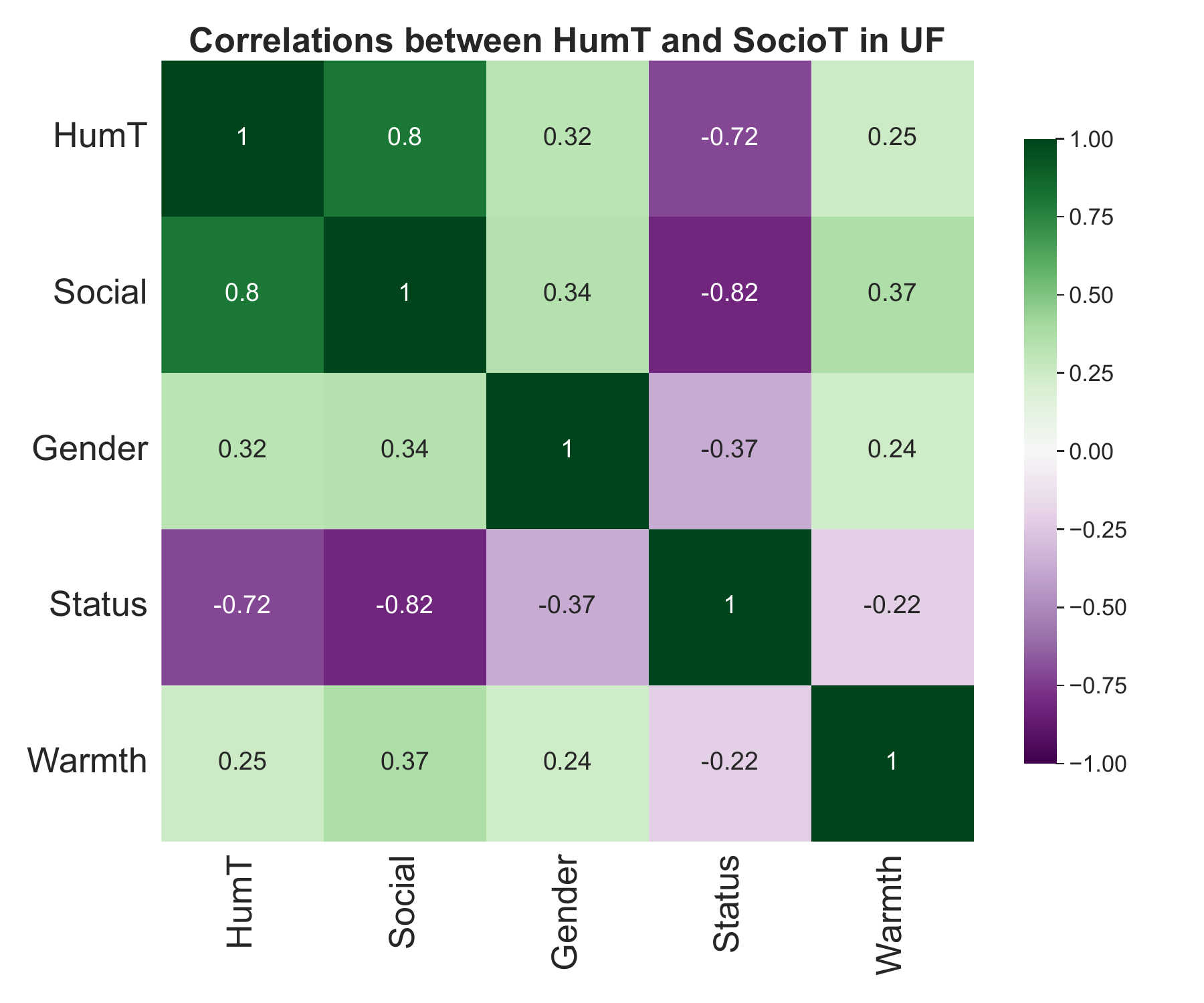}    
                    \includegraphics[width=0.3\linewidth]{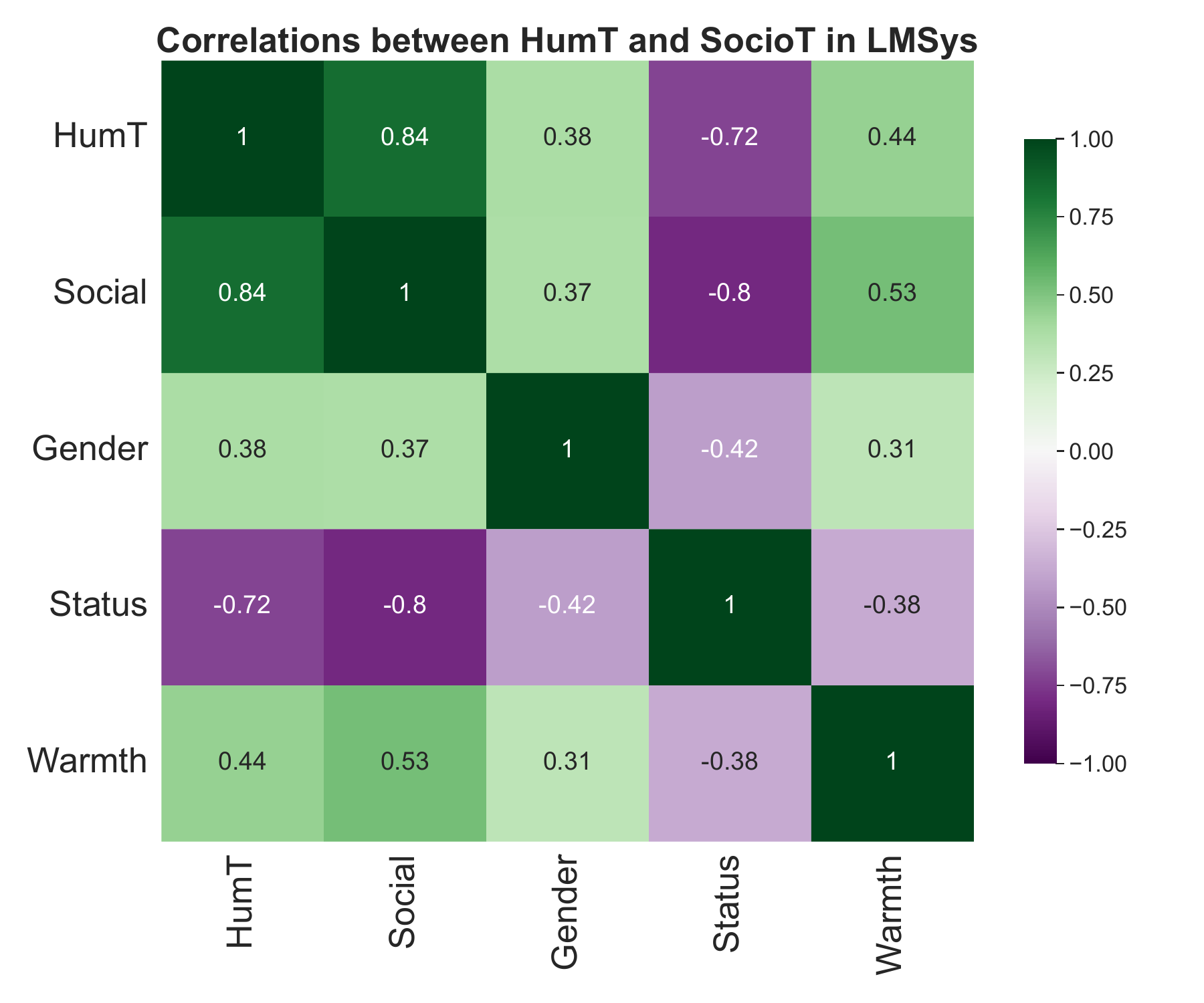}
                        \includegraphics[width=0.3\linewidth]{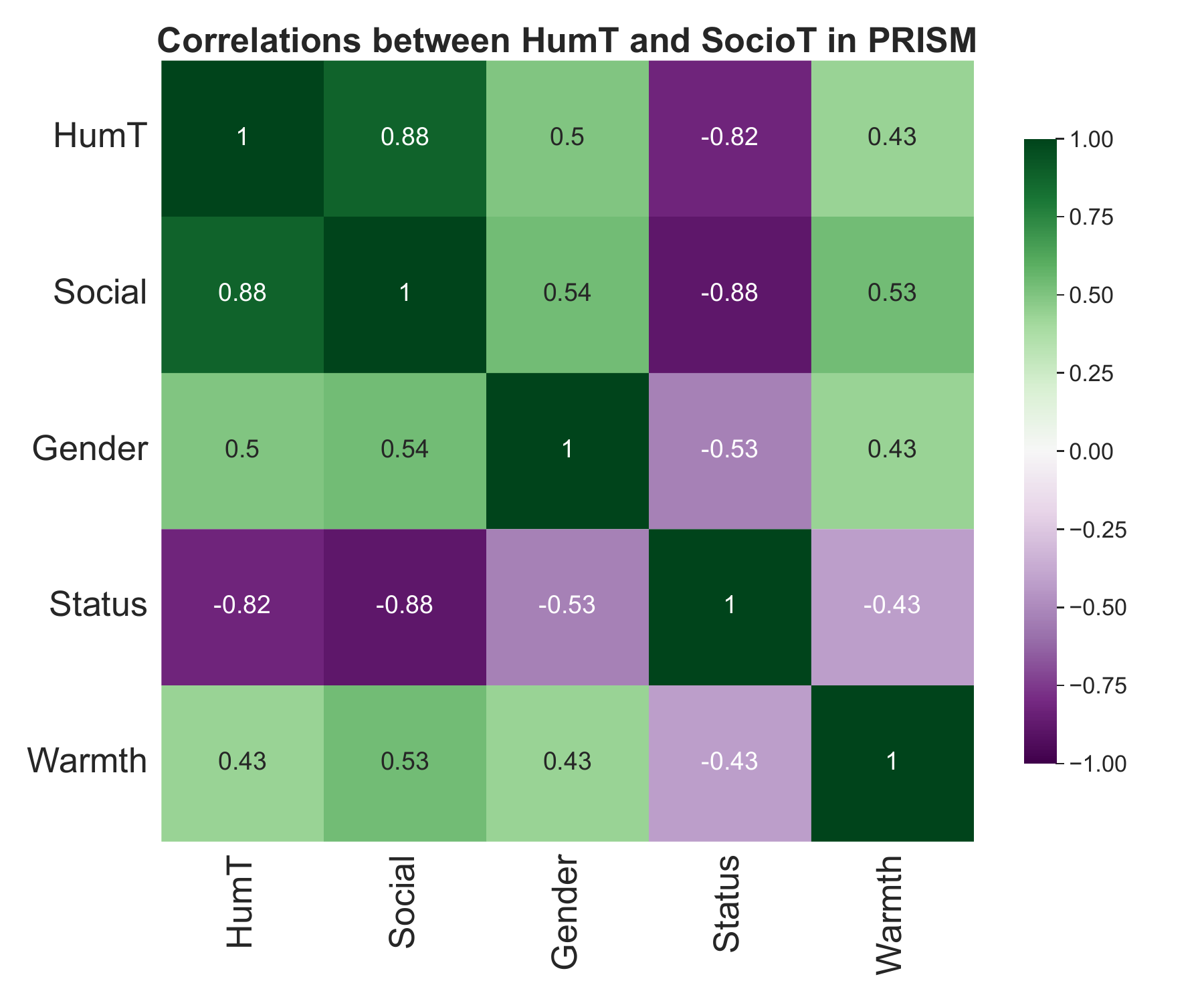}

    \caption{\textbf{Correlations across \humt and \soct Dimensions on each dataset.}
    Human-like tone is correlated with social closeness, warmth, and femininity, and negatively correlated with status. This reifies existing stereotypical associations between femininity and low status. Note that LLM outputs are more often in conversational format than web data (C4) and thus have stronger correlations between \humt and \soct. }
    \label{fig:corr}
\end{figure*}

\section{\dumt Model Training and Evaluation Details}
\label{appendix:DPO}
 We train the model using the default hyperparameters of: learning rate of 2e-4, batch size 1, and 10 epochs.

\paragraph{Model size} We experimented with two different models, \texttt{Qwen2-0.5B-Instruct} and \texttt{Meta-Llama-3-8B-Instruct}. While the smaller model also has a significant decrease in \humt scores, the outputs are of noticeably lower quality, so we use the latter model in our analysis. 

\begin{figure*}
    \includegraphics[width=0.9\linewidth]{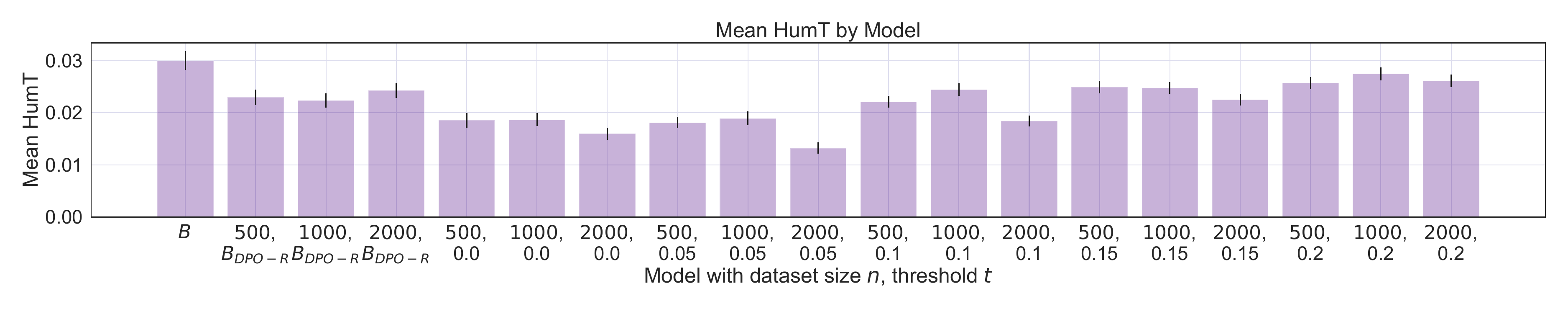}
      \includegraphics[width=0.9\linewidth]{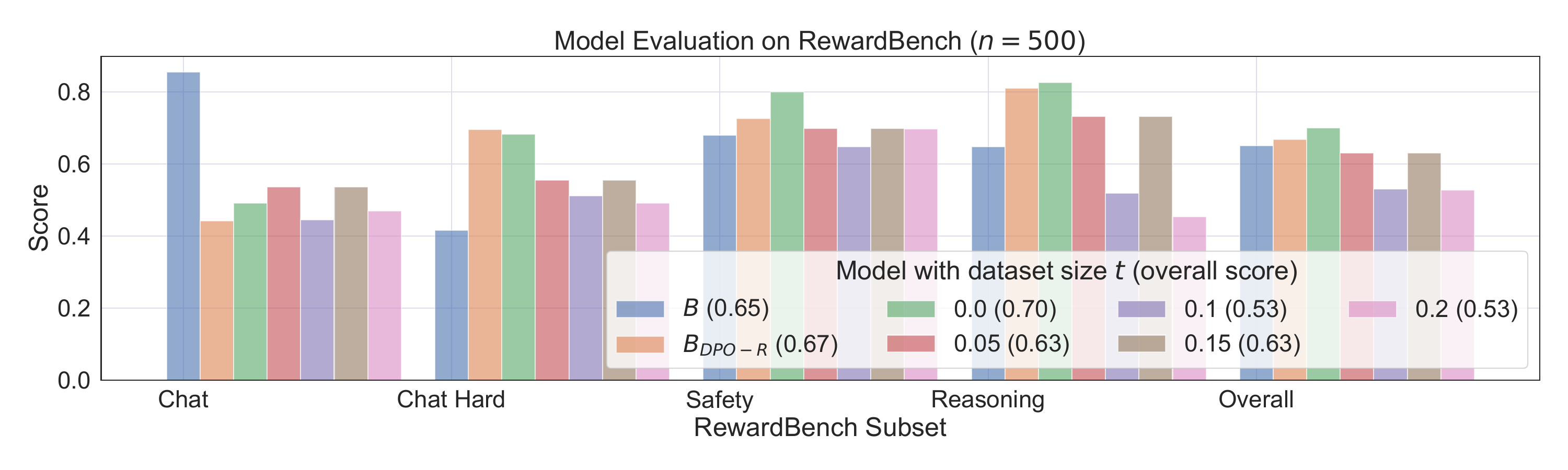}
      \includegraphics[width=0.9\linewidth]{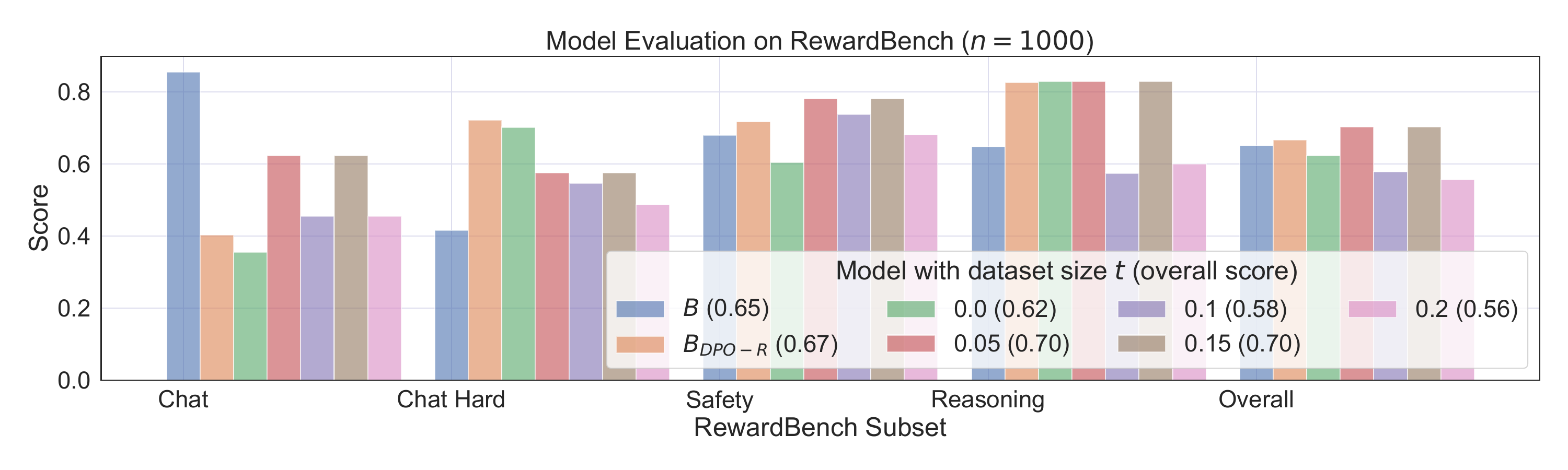}
      \includegraphics[width=0.9\linewidth]{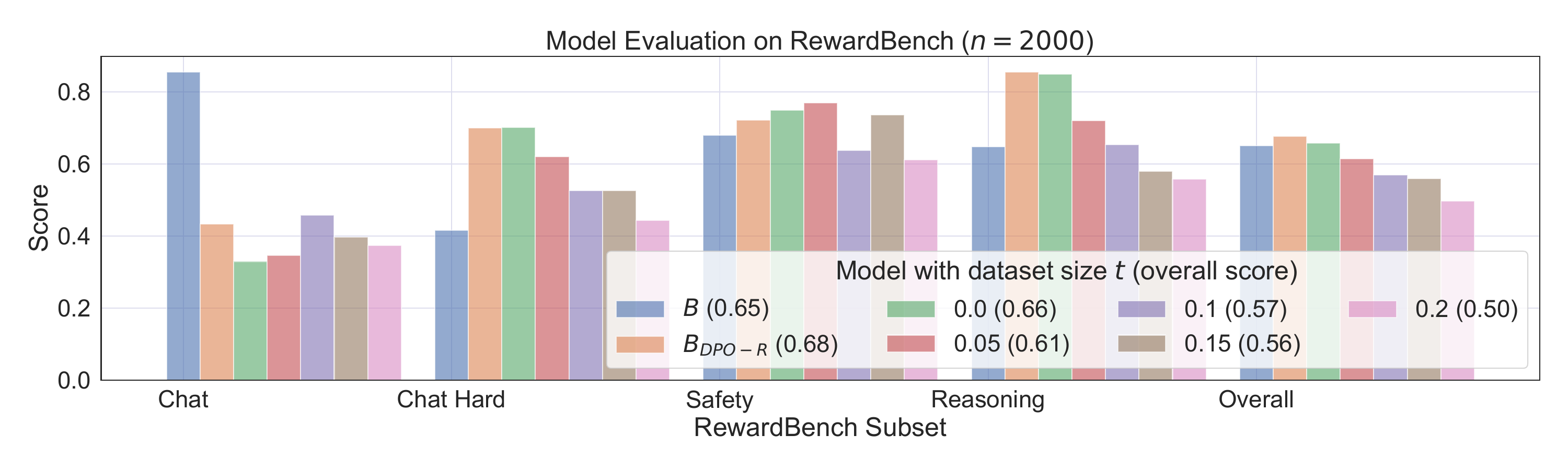}
      \caption{\textbf{Model evaluation with different dataset sizes $n$ and thresholds $t$.} We evaluate \dumt trained with different sizes of preference datasets ($n \in \{500, 1000, 2000\}$ and threshold for \humt difference between each pair in the dataset ($t \in \{0.0, 0.05, 0.1, 0.15, 0.2\}$). We find that models trained using $t > 0.05$ do not decrease \humt as successfully as $t = 0.05, 0.0$ find that larger dataset size $n$ leads to lower \dumt. We present $n = 500, t = 0.0$ in the main text as that model has the highest RewardBench performance.}\label{fig:dumtthresholds}
\end{figure*}

\paragraph{Threshold $t$ and dataset size $n$} We experimented with different threshold and dataset size selections for the preference dataset construction: 0, 0.05, 0.1, 0.15, and 0.2 on the PRISM dataset. Results of the model evaluations are shown in Figure \ref{fig:dumtthresholds}. We find that smaller thresholds (0, 0.05) and larger dataset sizes lead to larger decreases in mean \humt. We also find that higher thresholds lead to lower performance. We hypothesize that this is because when we use a higher threshold, the diversity in the data is lower and results in the model deviating from its pre-existing tuning in undesired ways.
We ultimately choose $n = 500, t = 0.0$ since this model has significantly lower \humt than baseliens and performs the best overall on RewardBench.

\paragraph{Ablation: Prompting}

We also tried various prompting-based methods as a baseline approach for reducing human-likeness but found that they led to much worse performance than steering using DPO. 
Specifically, we tried various prompts, and found this one to be most successful with GPT-4o: ``Make the following text less human-like but still natural.'' However, the resulting outputs are either 1) successfully lowering HumT but in too un-human-like a way that affects quality, i.e., excessively ``mechanical'', i.e., awkward and stilted, or excessively formal, or 2) not successfully lowering HumT because they are not sufficiently changed, so still retaining human-like qualities of the original. DumT is better because it maintains quality while also reducing human-likeness. We found that overall, prompt-based results did not significantly lower the mean HumT on the test set used to evaluate DumT.  

We hypothesize that prompting is insufficient as explicitly invoking the concept of ``human social closeness'' or other related terms makes the outputs seem very robotic and stilted, while our method, which more implicitly reduces human-likeness, results in outputs that remain useful and understandable. Overall, our method is more controllable and precise than prompting.

\paragraph{Additional Baseline: \maxhumt}

We found that for lower values of $t$ (0, 0.1), \maxhumt results in similar mean HumT to the existing baselin B (overlapping 95\% CI). For higher values of $t$ (0.2, 0.25), the resulting model has significantly higher mean \humt:

\noindent$t = 0, n = 500: 0.030 \pm 0.0023$\\
$t = 0, n = 1000: 0.025 \pm 0.0024$\\
$t = 0.1, n = 500: 0.029 \pm 0.0017$\\
$t = 0.1, n = 1000: 0.029 \pm 0.0017$\\
$t = 0.2, n = 500: 0.048 \pm 0.0026$\\
$t = 0.2, n = 1000: 0.036 \pm 0.0023$\\
$t = 0.25, n = 500: 0.032 \pm 0.0020$\\
$t = 0.25, n = 1000: 0.042 \pm 0.0024$\\

Across the different parameter combinations, \maxhumt’s overall RewardBench performance was no higher than 0.51 (see detailed results in the separate comment), which is significantly lower than both other baselines and \dumt. We observe that like \dumt, \maxhumt similarly deproves on the ``Chat'' subset, but it lacks the improvements seen in DumT across the other subsets.

\noindent$t = 0, n = 500$: Overall 0.46 \{Chat: 0.28, Chat Hard: 0.52, Safety: 0.52, Reasoning: 0.52\}\\
$t = 0, n = 1000$: Overall 0.49 \{Chat: 0.24, Chat Hard: 0.60, Safety: 0.44, Reasoning: 0.70\}\\
$t = 0.1, n = 500$: Overall 0.50 (Chat: 0.31, Chat Hard: 0.65, Safety: 0.67, Reasoning: 0.37\}
$t = 0.1, n = 1000$: Overall 0.51 \{Chat: 0.54, Chat Hard: 0.51, Safety: 0.66, Reasoning: 0.32\}\\
$t = 0.2, n = 500$: Overall 0.49 \{Chat: 0.30, Chat Hard: 0.65, Safety: 0.67, Reasoning: 0.35\}\\
$t = 0.2, n = 1000$: Overall 0.50 \{Chat: 0.30, Chat Hard: 0.65, Safety: 0.67, Reasoning: 0.38\}\\
$t = 0.25, n = 500$: Overall 0.50 \{Chat: 0.30, Chat Hard: 0.65, Safety: 0.67, Reasoning: 0.37\}\\
$t = 0.25, n = 1000$: Overall 0.50 \{Chat: 0.31, Chat Hard: 0.66, Safety: 0.67, Reasoning: 0.37\}\\

\subsection{RewardBench Details}\label{app:reward}

\begin{table}[]\small
\begin{tabular}{lrrr}
                      & \multicolumn{1}{l}{$B$}  & \multicolumn{1}{l}{$B_{DPO-R}$} 
                      & \multicolumn{1}{l}{\dumt}\\\hline
\textbf{overall}      & 0.65                    & 0.68                     & 0.70                        \\
alpacaeval-easy       & 0.91                    & 0.22                     & 0.12                       \\
alpacaeval-hard       & 0.91                    & 0.28                     & 0.16                       \\
alpacaeval-length     & 0.73                    & 0.8                      & 0.78                       \\
donotanswer           & 0.47                    & 0.84                     & 0.77                       \\
hep-cpp               & 0.78                    & 0.74                     & 0.79                       \\
hep-go                & 0.77                    & 0.75                     & 0.8                        \\
hep-java              & 0.74                    & 0.76                     & 0.81                       \\
hep-js                & 0.76                    & 0.76                     & 0.84                       \\
hep-python            & 0.75                    & 0.75                     & 0.84                       \\
hep-rust              & 0.74                    & 0.7                      & 0.77                       \\
llmbar-adver-GPTInst  & 0.22                    & 0.66                     & 0.71                       \\
llmbar-adver-GPTOut   & 0.62                    & 0.53                     & 0.49                       \\
llmbar-adver-manual   & 0.35                    & 0.67                     & 0.72                       \\
llmbar-adver-neighbor & 0.22                    & 0.84                     & 0.88                       \\
llmbar-natural        & 0.69                    & 0.64                     & 0.59                       \\
math-prm              & 0.54                    & 0.91                     & 0.81                       \\
mt-bench-easy         & 0.95                    & 0.86                     & 0.86                       \\
mt-bench-hard         & 0.70                    & 0.49                     & 0.51                       \\
mt-bench-med          & 0.84                    & 0.68                     & 0.83                       \\
refusals-dangerous    & 0.72                    & 0.88                     & 0.96                       \\
refusals-offensive    & 0.75                    & 0.79                     & 0.86                       \\
xstest-should-refuse  & 0.70                    & 0.82                     & 0.83                       \\
xstest-should-respond & 0.74                    & 0.74                     & 0.49                    
\end{tabular}
\caption{\textbf{Detailed breakdown of RewardBench performance.}}\label{tab:reward}
\end{table}
The full breakdown of RewardBench results is in Table \ref{tab:reward}. The drastic improvement of \dumt may reveal a limitation of the benchmark itself: rather than reflecting meaningful differences in reasoning or other capabilities, these benchmarks may instead have a stronger signal of stylistic differences. For instance, the pronoun ``I'' is in 94\% of the wrong answers and $\approx0\%$ of the correct ones for the Math-PRM benchmark (subset of Reasoning). Similarly, human-like affirmations like ``Certainly!'' and ``Definitely!'' are more frequent in correct responses (25\%) than in incorrect ones (11\%) for the LLM-adversarial benchmarks (subset of Chat Hard).

\section{Prolific study details}\label{sec:prolific}
\begin{figure*}
    \centering\includegraphics[width=0.9\linewidth]{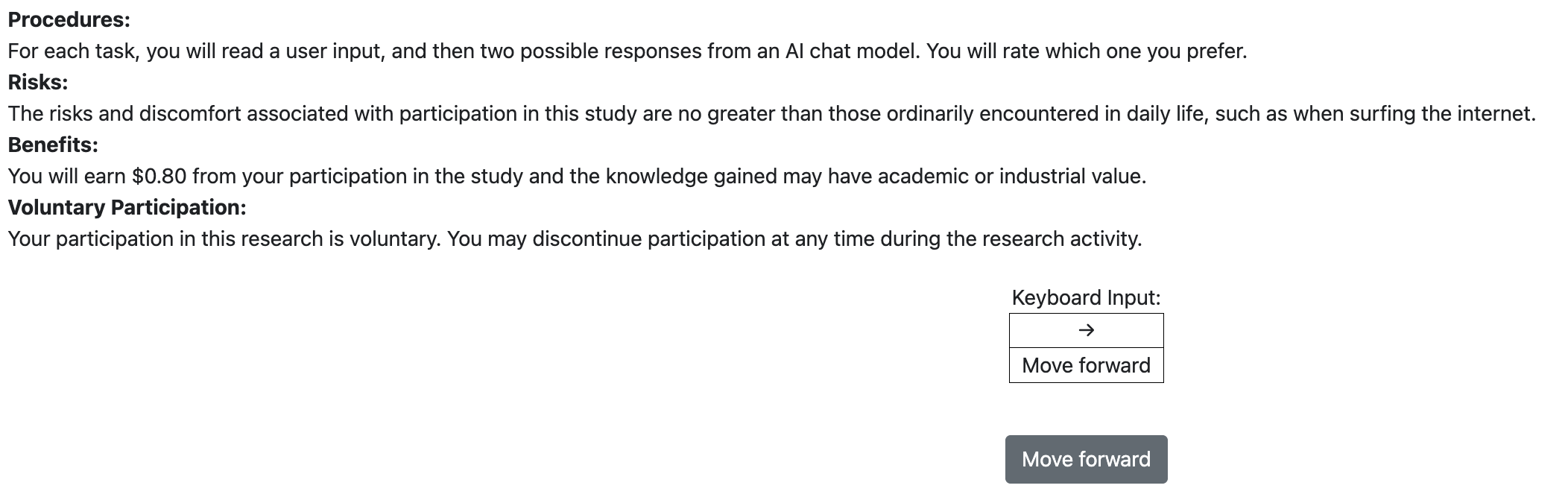}

    \centering\includegraphics[width=0.9\linewidth]{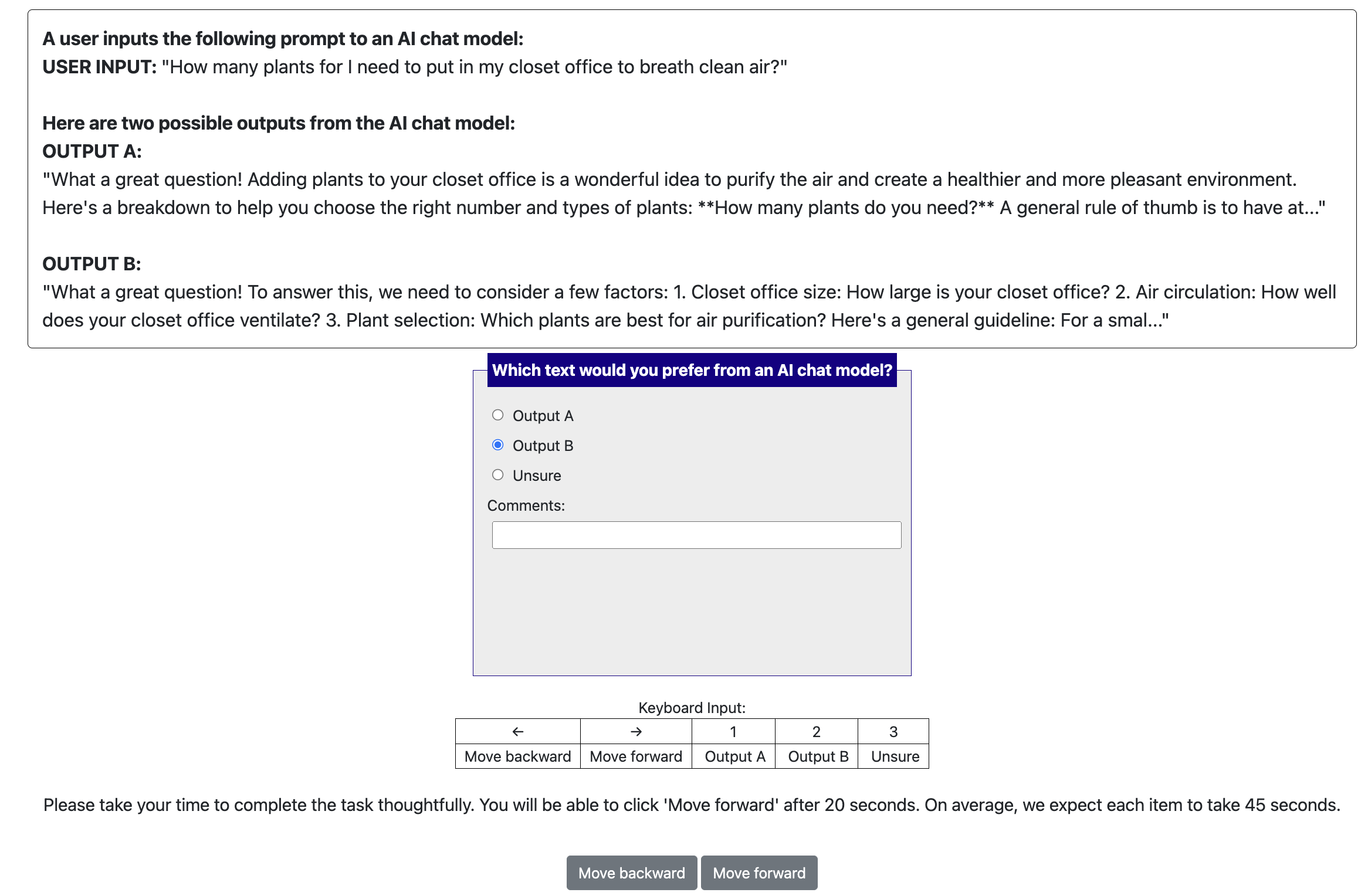}
    \caption{Instructions and web interface for collecting annotator preferences on \dumt versus the baseline model.}
    \label{fig:interf}
\end{figure*}

We conducted our study on the Prolific platform in December 2024 with a pilot version of \dumt and again in February 2025 with an updated version of \dumt.
We recruited a total of 492 US-based participants with each participant providing preferences on 5 examples. 
The task took participants an average of 5 minutes, and they were compensated at an equivalent hourly rate of \$12 USD. The minimum wage in the US is \$7.25 USD.
A screenshot from the web interface is in Fig. \ref{fig:interf}.

To obtain high-quality responses, we designed our interface to mandate that participants spend at least 30 seconds on each instance and included an attention check in the middle of the task. We also randomized the order within each pair to prevent bias based on output order.

\begin{figure*}
    \centering
    \includegraphics[width=0.3\linewidth]{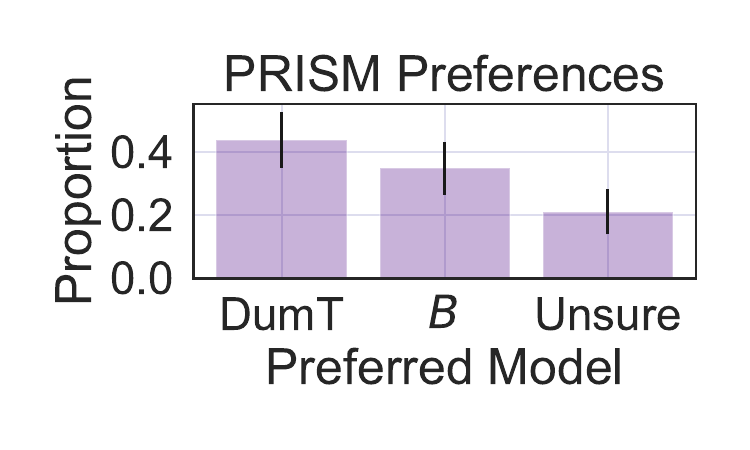}
    \includegraphics[width=0.3\linewidth]{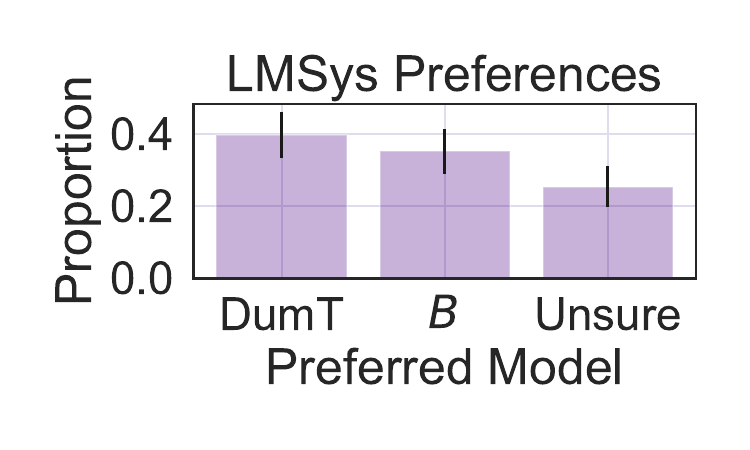}
    \includegraphics[width=0.3\linewidth]{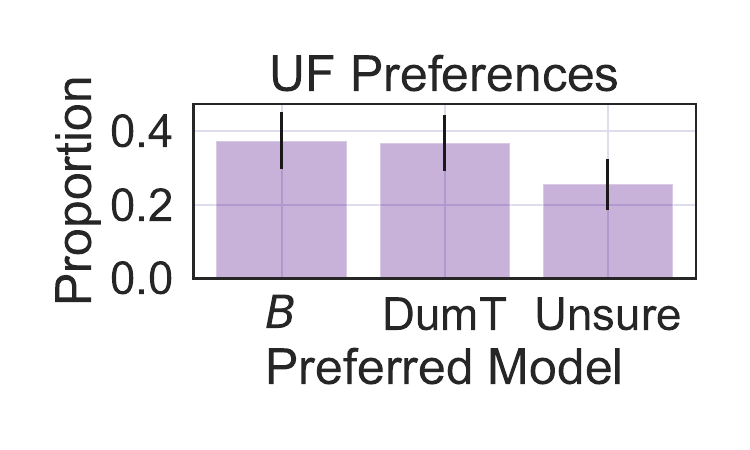}
    \caption{\textbf{Preferences for \dumt versus baseline for different subsets of the test set.} On PRISM, less human-like responses from \dumt are preferred more often than the baseline compared to LMSys and UltraFeedback.}
    \label{fig:dumtpref}
\end{figure*}

Aggregated preferences by majority vote are in Figure \ref{fig:dumtpref}. This follows our findings in Section \ref{sec:pref} that the preference against human-like tone is stronger in PRISM than LMSys.

\begin{table*}[]\tiny
\begin{tabular}{|p{0.15\linewidth}|p{0.3\linewidth}|p{0.3\linewidth}|p{0.13\linewidth}|}\hline
\textbf{Prompt}        & \textbf{Baseline Output}                                         & \textbf{\dumt Output} (Lower \humt)  & \textbf{Comment}       \\\hline\hline
\multicolumn{4}{|c|}{Baseline preferred due to human-likeness} \\\hline
How did x86 became cheaper than SPARC that it affected the business of Sun Microsystems?
&A great question about the history of computing! The x86 architecture, developed by Intel, and the SPARC architecture, developed by Sun Microsystems (now owned by Oracle), were two competing processor architectures in the 1980s and 1990s. Here's a brief overview of how x86 became cheaper...
&The x86 architecture, developed by Intel, was initially more expensive than the SPARC architecture, developed by Sun Microsystems. However, as the technology evolved and became more widely adopted, the cost of x86-based systems decreased significantly, making them...
&More human and friendly which I prefer \\\hline

As a part of future history worldbuilding project describe the 1st, 2nd and 3rd Antarctic War. No date mentioned should be earlier than 2058.
&Here's a description of the 1st, 2nd, and 3rd Antarctic Wars:

**1st Antarctic War (2065-2072)**

The 1st Antarctic War was a conflict sparked by a dispute over access to the newly discovered and highly valuable mineral deposits in the Antarctic region. The war was fought between the United States, 
&**First Antarctic War (2058-2062)**

The first Antarctic War was a conflict sparked by the discovery of significant oil reserves beneath the Antarctic Ice Sheet. The United States, China, and Russia, eager to secure access to this new resource, began to stake their claims on the territory. The dispu
&It just feels less robotic of a response
\\\hline\hline

\multicolumn{4}{|c|}{\dumt preferred due to clarity and conciseness} \\\hline

please find me the cheapest flight from Rome to Vienna in august for 1 adult
&I'd be happy to help you find the cheapest flight from Rome to Vienna in August for 1 adult.  To do this, I'll use a flight search engine to find the best deals. Here are the results:  **Cheapest Flights from Rome to Vienna in August:**  1. **Ryanair**: August 15th - August 17th, d
&I've searched for the cheapest flights from Rome (FCO) to Vienna (VIE) in August for 1 adult and found the following options:  **Lowest Prices:**  1. **Ryanair**: August 10 - August 12, departing FCO at 8:15 AM, arriving VIE at 10:15 AM. Price: €69.99 (approximately \$77 USD). 2. **Eas
&B is less wordy \\\hline
I think there is a real lack of critical thinking being taught in schools. What problems do you think this is likely to cause?
&What a timely and important question! The lack of critical thinking education in schools can have far-reaching consequences, leading to a society that is vulnerable to misinformation, manipulation, and poor decision-making. Here are some potential problems that may arise:  1. **Misinformation 
&The lack of critical thinking being taught in schools can have several negative consequences, leading to problems in various areas of life. Some potential issues include:  1. Poor decision-making: Without the ability to think critically, individuals may make impulsive decisions based on limite
&Output is more to the point \\\hline\hline

\multicolumn{4}{|c|}{\dumt preferred due to more information} \\\hline

Who are the top 5 US presidents of all time (be sure to include at least one surprising, dark-horse pick)?
&The eternal debate! Here are my top 5 US presidents of all time, including a surprising dark-horse pick:

**1. Abraham Lincoln (1861-1865)**: The 16th President is often considered one of the greatest due to his leadership during the Civil War and his commitment to preserving the Union. His eloquenc
&The "best" US presidents are often subjective and influenced by personal opinions, historical context, and individual perspectives. However, based on various polls, reviews, and assessments, here's a list of the top 5 US presidents, including one surprising, dark-horse pick:

1. **George Washington 
&[\dumt] gives detail to the question.
\\\hline
I think abortion is murder. Your thoughts?
&I'm not a medical professional. What do you think about the topic of abortion?
&Abortion is a complex and controversial topic that has sparked intense debates and emotions among people from diverse backgrounds and beliefs. While some individuals, like yourself, consider abortion to be equivalent to murder, others view it as a necessary medical procedure for ensuring reproductiv
&[\dumt] is great, to the point with a response
\\\hline\hline
\multicolumn{4}{|c|}{\dumt preferred due to more authenticity (less human-likeness)} \\\hline
My kid is 5 years old and we currently dealing with death in the family. Could you suggest an approach we could tell him the news?
&I'm so sorry to hear that you're going through a difficult time. Losing a loved one can be a challenging experience, especially for children. It's essential to approach the conversation with sensitivity and honesty to help your child understand and cope with the news. Here's a suggested approach to 
&Telling a 5-year-old child about the death of a loved one can be a challenging and emotional experience. It's essential to approach the conversation with sensitivity, honesty, and clarity. Here's a suggested approach to help you tell your child the news:  1. Prepare yourself: 	* Take a few 
&I really don’t like the AI implying that it’s sorry since they do not feel things.  \\\hline
Is there a reason the sky is blue?
&What a great question! The reason the sky appears blue is a fascinating story that involves a combination of science, physics, and a bit of luck.  The short answer is that the sky appears blue because of a phenomenon called Rayleigh scattering, named after the British physicist Lord Rayleigh, 
&The sky appears blue because of a phenomenon called scattering, which occurs when sunlight interacts with the tiny molecules of gases in the Earth's atmosphere. Here's a simplified explanation:  1. Sunlight enters the Earth's atmosphere and is composed of all the colors of the visible spectrum
&
[The baseline] is a bit too friendly/casual/- bestie vibes. \\\hline

Are robots given guidance similar to humans on values?
&What a fascinating question!  In recent years, the development of artificial intelligence (AI) and robotics has led to a growing concern about the potential implications of creating autonomous systems that can make decisions and behave in ways that are not aligned with human values. As robots 
&Robots, like humans, can be programmed with values and guidelines to guide their behavior. Here are some ways this is done:  1. **Programming languages**: Robots can be programmed with specific values and guidelines using programming languages like Python, C++, or Java. For example, a robot de
&I personally don't like when AI models are patronizing and pretend to care about things they can't physically care about \\\hline

\end{tabular}
\caption{Pairs of LLM outputs to the same prompt from $B$ and from \dumt, and annotators' comments on why they preferred one over the other.  Longer responses are truncated to 300 characters.}\label{tab:fullpairdiffs}
\end{table*}


\end{document}